\documentclass[11pt,a4paper,oneside]{article}
\usepackage[english]{babel}

\usepackage[T1]{fontenc}
\usepackage{kpfonts}

\usepackage{geometry}
\usepackage[activate={true,nocompatibility}]{microtype}

\usepackage{tabularx}
\usepackage{booktabs}
\usepackage{makecell}

\usepackage[style=authoryear-ibid,
isbn=false,
url=false,
eprint=false,
date=year,
maxcitenames=2,
mincitenames=1,
maxbibnames=999,
minbibnames=1,
backend=biber]{biblatex}
\usepackage{hyperref}
\usepackage{doi}
\hypersetup{
	hidelinks
}

\usepackage{csquotes}
\usepackage{lettrine}
\usepackage{booktabs}
\usepackage{dcolumn}

\usepackage{graphicx}
\usepackage{subcaption}

\usepackage{caption}
\title{Measuring Negative Campaigning across Languages with Large Language Models: A Study of 18 Million Tweets in 19 Countries}

\author{
Victor Hartman \\
\small University of Amsterdam
\and
Petter T\"ornberg\thanks{Corresponding author: p.tornberg@uva.nl} \\
\small University of Amsterdam
}

\date{\today}

\begin{document}
	
\maketitle

\begin{abstract}\noindent Negative campaigning is a defining feature of electoral competition, yet comparative research on its drivers has remained limited by the high cost and limited scalability of
existing classification methods. This study makes two key contributions. First, it evaluates zero-shot large language models (LLMs) as a scalable method for cross-lingual classification of negative campaigning. Using benchmark datasets in ten languages, we show that LLM classifications closely match native-speaker human annotations while outperforming conventional supervised models. Second, we leverage this approach to conduct, to our knowledge, the largest cross-national study of negative campaigning to date, analyzing 18 million tweets posted by parliamentarians in 19 European countries between 2017 and 2022. Building on a strategic incentives framework, we argue that governing and coalition-oriented parties face stronger reputational constraints against negative campaigning, whereas opposition and outsider parties face weaker constraints. We further expect parties located away from the ideological center, especially on the radical right, to rely more heavily on confrontational rhetoric. The results support these expectations: cabinet parties are less negative, while ideologically non-centrist parties—most notably radical right parties—are substantially more negative. The study thus provides new comparative evidence on the party-level foundations of campaign negativity while demonstrating how LLMs can transform the scale, consistency, and replicability of research on political discourse across languages and institutional contexts.
\end{abstract}


\vspace{1em}

\noindent Negative campaigning---commonly defined as ``any criticism levelled by
one candidate against another during a campaign'' \parencite[23]{geerDefenseNegativityAttack2006}---has long been a feature of democratic politics. In ancient Athens, organized efforts to exile political rivals through ostracism were inscribed on pottery shards used as ballots \parencite{brenneOstrakaVomKerameikos2020}. Yet recent scholarship has raised growing concern that negativity in campaigning may be increasing, particularly as political communication adapts to the affordances and logics of social media \parencite{auterNegativeCampaigningSocial2016,klingerAreCampaignsGetting2023,tornbergWhenPartiesLie2025}. These concerns are not merely stylistic: negative campaigning has been linked to lower trust in political institutions \parencite{lauEffectsNegativePolitical2007} and to rising affective polarization \parencite{lauEffectMediaEnvironment2017,martinDeepeningRiftNegative2024}.

Yet the political drivers of negative campaigning remain contested. Much existing research has focused on candidate-level factors such as personal characteristics and electoral competitiveness, largely in single-country settings and especially in the United States. While this literature has produced important insights, it is also shaped by assumptions that fit majoritarian, candidate-centered systems better than multiparty democracies, where parties are the key strategic actors and where coalition incentives structure electoral competition. As a result, party-level explanations—particularly those concerning governing status, ideological positioning, and related coalition dynamics—have received comparatively less systematic attention in large-scale cross-national analyses based on direct behavioral data \parencite{haselmayer2019negative,valliNai2020}.

This gap is due in part to methodological constraints. Like many phenomena in political communication, negative campaigning is difficult to study comparatively at scale because it is a subtle and context-dependent phenomenon. Existing research has relied primarily on either expert surveys or manual content analysis. Expert-based approaches are relatively feasible in comparative settings, but they raise familiar concerns about bias, limited transparency, and weak replicability across contexts \parencite{lindstadtWhenExpertsDisagree2020}. Manual coding offers closer grounding to observed messages, but it is labor-intensive, difficult to scale, and often infeasible for multilingual cross-country datasets. Automated approaches based on supervised machine learning have expanded the scale of analysis \parencite{petkevicPoliticalAttacks2802022,lichtMeasuringUnderstandingParties2024}, but they remain constrained by their dependence on extensive labeled training data for each language and context, and often underperform relative to human coding on subtle interpretive tasks \parencite{vanatteveldtValiditySentimentAnalysis2021}. Together, these limitations have made it difficult to generate direct, comparable, and fine-grained measures of negative campaigning across countries.

Recent advances in large language models (LLMs) offer a potential way forward. Instruction-tuned LLMs enable ``zero-shot'' classification, allowing researchers to specify coding instructions in natural language without requiring task-specific training data \parencite{kojimaLargeLanguageModels2023}. Early research suggests that LLMs can perform classification tasks across domains and languages with high accuracy and at dramatically lower cost than traditional annotation pipelines \parencite{gilardiChatGPTOutperformsCrowd2023,tornbergLargeLanguageModels2024}. Unlike conventional supervised models, which are typically trained on relatively small and domain-specific datasets, LLMs are trained on large and diverse text corpora and can draw on broad linguistic and contextual knowledge. This gives them the potential to generalize across languages, political settings, and subtle semantic distinctions. Scholars have therefore begun to explore whether LLMs can enable more consistent annotation across languages, with important implications for comparative political communication research \parencite{rathjeGPTEffectiveTool2024,tornbergLargeLanguageModels2024}. Yet systematic evidence on whether LLMs can produce high-quality cross-lingual data for nuanced political concepts such as negative campaigning remains limited.

This paper addresses that gap. To our knowledge, it provides the first systematic evaluation of zero-shot LLM annotation of negative campaigning across multiple languages and country contexts. We assess model performance against two high-quality manually coded datasets: Petkevic and Nai's (2022) study of the 2018 U.S. Senate elections and Klinger et al.'s (2023) multilingual coding of the 2014 and 2019 European Parliament campaigns. We treat these datasets as established benchmarks for measurement validity. Comparing LLM classifications to these human-coded annotations allows us to assess the extent to which automated classification captures the underlying concept of negative campaigning. The results show that zero-shot LLMs achieve performance comparable to, and in some cases exceeding, the available human-coded benchmarks across ten languages (English, German, Croatian, French, Hungarian, Italian, Dutch, Polish, Spanish, and Swedish), indicating that they can support valid and scalable cross-national analysis.

We then leverage this method to conduct a large-scale cross-national study of negative campaigning. Although existing research has generated important insights, it has remained disproportionately focused on single-country contexts---especially the United States \parencite[see][for overviews]{haselmayer2019negative,naiWalter2015}---while large-scale comparative analyses using direct behavioral measures remain rare, particularly in multilingual settings \parencite{valliNai2020,maierNaiVerhaar2024}. As a result, our understanding of negative campaigning in multiparty systems remains incomplete. Prior research has primarily examined macro-level contextual drivers \parencite{pappMacroLevelDrivingFactors2019,maierWhenConflictFuels2022} or candidate-level determinants \parencite{maierMappingDriversNegative2023}. And while some comparative party-level research exists \parencite{walterWhenGlovesCome2013,walterWhenStakesAre2014,baranowski2024patterns}, large-scale cross-national studies that systematically link party characteristics---such as ideology, governing status, or populism---to negative campaign behavior using direct behavioral measures remain scarce. This is a consequential gap, since parties---rather than individual candidates---are the central strategic actors in many proportional and coalition-based systems. Moreover, much of the existing comparative literature relies on expert surveys or self-reported behavior, which can limit both the granularity and the measurement validity of negativity indicators \parencite{lindstadtWhenExpertsDisagree2020}.

Our empirical analysis contributes by focusing on parties as the unit of analysis and by relying on direct observations of their textual communication. Building on a strategic incentives framework, we develop expectations about how party characteristics shape the use of negative campaigning in multiparty contexts. We test these expectations using a dataset of 18 million messages posted by elected representatives across 19 European countries. This design allows us to examine how cabinet participation, ideological positioning, proximity to elections, and populist discourse are associated with the use of negative campaigning in observed communication, thereby linking theories of party competition to systematic patterns of political messaging.

Our findings show that cabinet participation is consistently associated with lower levels of negative campaigning. We also find evidence that negativity tends to be higher among parties located farther from the ideological center. Across the party-family analysis, radical right parties stand out as the most negative, with radical left parties also showing elevated levels. By contrast, the positive associations of right-wing positioning, populism, and election proximity are weaker and depend more on model specification and sample.
Taken together, these findings suggest that rising negativity in political communication \parencite{klingerAreCampaignsGetting2023,tornbergWhenPartiesLie2025} is not simply a byproduct of social media logics, but also reflects deeper dynamics of party competition, coalition incentives, and the growing prominence of particular political movements.

\section{Measuring Negative Campaigning}\label{measuring-negative-campaigning}

Negative campaigning is traditionally measured using either expert
surveys or manual annotation of political messaging. Both approaches,
however, come with well-documented limitations.

Expert surveys have been widely used to measure negative campaigning in
comparative research due to their relatively low cost and feasibility
across multiple countries \parencite[e.g.,][]{naiWalter2015}. A key
strength of this approach is its ability to capture holistic, latent
judgments about the overall tenor of a campaign---assessments that may
extend beyond what is observable in individual messages alone. However,
the approach also has several well-documented limitations. First, expert
evaluations are inherently subjective, introducing potential biases
based on the experts' political preferences, media exposure, or national
context \parencite{lindstadtWhenExpertsDisagree2020, walterUnintendedConsequencesNegative2019}. These
biases can be particularly problematic in cross-national settings, where
cultural and linguistic differences may influence what is perceived as
``negative'' campaigning \parencite{esserComparingNewsNational2012, esserComparingPoliticalCommunication2004}. Second, expert assessments often lack transparency,
as the criteria used by individual raters are rarely standardized or
externally validated, raising concerns about the reliability and
replicability of the data \parencite{mikhaylovCoderReliabilityMisclassification2012, volkensStrengthsWeaknessesApproaches2007}. Third, the aggregated nature of expert surveys tends to obscure
the granularity of campaign messaging, failing to capture variation in
tone across messages, platforms, or time \parencite{vanatteveldtValiditySentimentAnalysis2021}.
This makes it difficult to analyze how negativity evolves in response to
political events or strategic shifts. Collectively, these limitations
suggest that while expert surveys offer valuable broad-level assessments, they fall short
in capturing the nuance and context-dependent nature of negative
campaigning, particularly in multilingual and multi-platform
environments.

In contrast to expert surveys, manually coding political messages
offers several methodological advantages for
measuring negative campaigning \parencite[e.g.][]{geerDefenseNegativityAttack2006, lauNegativeCampaigningAnalysis2004a, auterNegativeCampaigningSocial2016, fridkinVariabilityCitizensReactions2011, elmelund-praestekaerAmericanNegativityGeneral2010, hansenNegativeCampaigningMultiparty2008, walterWhenGlovesCome2013}. Rather than
relying on second-order perceptions, manual coding evaluates the actual
content of campaign messages---such as tweets, speeches, or
advertisements---allowing researchers to systematically apply consistent
coding rules across cases. This improves both the validity and
replicability of findings, especially when coders are trained using a
shared codebook and inter-coder reliability is assessed \parencite{krippendorffContentAnalysisIntroduction2019}. Manual coding also enables a higher level of granularity,
capturing variation within parties, over time, or across
platforms---details that are typically obscured in aggregate expert
ratings \parencite{grimmerTextDataPromise2013}.

These advantages notwithstanding, manually coding political messages
comes with significant limitations of its own. High reliability and accuracy of manual
annotation are not guaranteed, even with extensive training \parencite{weberMeasuringSentimentPolitical2018}. To maximize classification accuracy, human coders typically
undergo training sessions where they develop a shared, often tacit
understanding of the concepts they are classifying---which often proves
difficult to document explicitly and cannot be fully captured in
codebooks \parencite{vanatteveldtValiditySentimentAnalysis2021}. This makes replication more difficult
and can leave the concept vaguely defined and variously
operationalized within the field, posing serious conceptual challenges.

Manual annotation is also highly resource-intensive: developing reliable
codebooks, training coders, and ensuring inter-coder consistency require
substantial time, labor, and funding \parencite{krippendorffContentAnalysisIntroduction2019, vanatteveldtValiditySentimentAnalysis2021}. While crowd-sourcing offers a partial solution by leveraging the
``wisdom of the crowd'' to scale annotation efforts \parencite{benoitCrowdsourcedTextAnalysis2016, vanatteveldtValiditySentimentAnalysis2021}, studies consistently find that trained
coders outperform crowd-workers in both accuracy and reliability \parencite{vanatteveldtValiditySentimentAnalysis2021}. Beyond performance differences, crowd-sourcing introduces
its own validity concerns, including variable coder motivation and
attention, which can compromise annotation quality in ways that are
difficult to detect or control for \parencite{benoitCrowdsourcedTextAnalysis2016}. Moreover, even crowd-based approaches remain too
costly for large-scale or multilingual analyses. The challenges are
particularly acute in cross-linguistic research, where consistency
across languages and cultural frames requires aligning coders from
different backgrounds around shared definitions---further increasing
complexity and cost \parencite{esserComparingNewsNational2012}.

As a result, most research on negative campaigning has relied on small
samples or single-country studies, limiting generalizability and
impeding broader comparative insights. The few comparative studies that
do exist illustrate both the demand for and difficulty of cross-national
analysis: \textcite{walterWhenStakesAre2014} study elections in the
Netherlands, UK, and Germany using manual coding but are necessarily
constrained in scale; \textcite{klingerAreCampaignsGetting2023} analyze
European Parliament campaigns but face similar scalability challenges;
and \textcite{naiGoingNegativeWorldwide2020} achieves broader geographic
coverage by relying on expert surveys rather than direct content
analysis, inheriting the limitations discussed above. These trade-offs
between depth and breadth underscore the need for measurement approaches
that can combine the granularity of content-level analysis with the
scalability required for comparative research.

These methodological issues have motivated efforts to automate the
classification of negativity. However, while conventional
dictionary-based approaches are fast and cost-efficient, they achieve
limited performance on semantically subtle concepts such as negative
campaigning, and perform even worse in multilingual settings \parencite{haselmayerSentimentAnalysisPolitical2017}. More recent studies have adopted more
sophisticated supervised machine learning techniques that leverage
semantic features and context. For instance, \textcite{petkevicPoliticalAttacks2802022} use
a multilayer perceptron to classify negativity in tweets, achieving a
macro F1-score of .82. \textcite{lichtMeasuringUnderstandingParties2024} employ a fine-tuned
transformer model to detect anti-elite sentiment, reaching an F1-score
of .75. \textcite{widmannCreatingComparingDictionary2023} use similar models to classify emotional
content, with F1-scores ranging from .60 (sadness, pride) to .84
(anger). These scores are relatively impressive given the difficulty of
the task and often modest agreement among human annotators on the
training data. Nevertheless, automated classifiers still lag behind
well-trained human coders in both accuracy and interpretive nuance \parencite{vanatteveldtValiditySentimentAnalysis2021}. More crucially, supervised machine learning still
depends on large-scale manually labeled training data for each specific
task and context. To build reliable models, researchers must first
produce extensive hand-coded datasets, often repeating the annotation
process for each new country, time period, or linguistic context \parencite{raffelExploringLimitsTransfer2023, petkevicPoliticalAttacks2802022, keucheniusWhyItImportant2021, widmannCreatingComparingDictionary2023, lichtMeasuringUnderstandingParties2024}. As a result, supervised machine
learning inherits many of the same limitations as manual annotation,
limiting its applicability for fully comparative, multilingual political
communication research.

The recent emergence of Large Language Models (LLMs) such as GPT has
however offered hopes of overcoming these challenges, having been described as representing a
paradigm shift in natural language processing \parencite{tornbergLargeLanguageModels2024}. Centrally, LLMs are capable of `few-shot' and `zero-shot' annotation: by using carefully
formulated natural language instructions, LLMs can be used to directly
annotate data without the need for manually labelled training data \parencite{bailCanGenerativeAI2024, brownLanguageModelsAre2020, kojimaLargeLanguageModels2023, tornbergLargeLanguageModels2024}.
By leveraging its understanding of language patterns gained through
training on vast textual data, the model can perform tasks it has not been
explicitly trained on \parencite{brownLanguageModelsAre2020}. Not only does this make the
models orders of magnitude cheaper, more flexible and easier to use, but
it may potentially have made them better annotators for many tasks: LLMs
have been shown capable of outperforming human coders, crowd-sourced coders,
experts and supervised machine learning methods in many classification
tasks \parencite{tornbergLargeLanguageModels2024, gilardiChatGPTOutperformsCrowd2023}.

LLMs also bring another capacity that potentially makes them
particularly impactful in the context of comparative research: scholars
have argued that their cross-linguistic capacities enable their use to
produce comparative data across languages and country contexts \parencite{rathjeGPTEffectiveTool2024, tornbergLargeLanguageModels2024}. This suggests that zero-shot prompting may enable consistent
classification across languages without requiring separate models or
training data for each language, while outperforming other methods in
accuracy.

However, scholars have argued that the performance of models varies
across languages, and the capacity to use LLMs to produce comparative
data has yet to be empirically investigated. The capacities of LLMs are
moreover fickle: while they achieve superhuman performance on some
tasks, they can seemingly inexplicably fail on other \parencite{ollionChatGPTTextAnnotation2023, kristensen-mclachlanAreChatbotsReliable2025, yuOpenClosedSmall2023}. Moreover,
LLMs can reproduce or amplify social and cultural biases embedded in
their training data, raising important ethical and methodological
concerns \parencite{lucyGenderRepresentationBias2021,tornberg2026largelanguagemodelsreproduce}. Scholars have therefore argued
that rigorous validation for each use case remains essential \parencite{tornbergBestPracticesText2024}.

\section{Method: Assessing LLMs' capacity to classify negative campaigning}

To evaluate the capacity of LLMs to accurately and reliably classify
negative campaigning across languages and country-contexts, we compare
the zero-shot LLM annotation against the manual classification data from
two studies: \textcite{petkevicPoliticalAttacks2802022}, covering Twitter posts of
candidates during the 2018 US Senate elections, and \textcite{klingerAreCampaignsGetting2023}, covering Facebook posts of political parties during the 2014 and
2019 EP election campaigns. These datasets are not publicly shared, which makes direct inclusion in LLM training data unlikely, although it cannot be fully ruled out given the limited transparency of training corpora. The
data are moreover of high quality, and offer coding by trained native
speakers across ten different languages and country-contexts (English,
German, Croatian, French, Hungarian, Italian, Dutch, Polish, Spanish,
and Swedish).

As is generally the case, these two studies use different definitions of
negative campaigning: \textcite{klingerAreCampaignsGetting2023} uses a significantly stricter
definition of negative campaigning than \textcite{petkevicPoliticalAttacks2802022}. This
difference also allows us to evaluate how well LLMs function across
different definitions (see Supplementary Material for details).

Following standard practice in machine learning, we evaluate model performance using F1 scores. The F1 score combines precision (the share of predicted positives that are correct) and recall (the share of actual positives that are correctly identified) into a single metric, providing a balanced measure of classification performance. This is particularly useful in settings with class imbalance, where raw accuracy can be misleading—for example, when one class is much more common than the other, a trivial classifier can achieve high accuracy by always predicting the majority class. 

In such contexts, macro-averaged F1 can also be misleading, as it assigns equal weight to each class regardless of their prevalence. We therefore report weighted F1 scores, which compute the average F1 across classes while weighting each class by its frequency in the data. This provides a more informative summary of overall performance in datasets with uneven class distributions, such as the \textcite{klingerAreCampaignsGetting2023} data. (See Supplementary Material for additional validation metric details, such as the confusion matrices.)

We moreover use inter-rater reliability (IRR) measures to assess the
degree of agreement between human coders and the LLM, treating the LLM
as another rater to benchmark against human performance standards.
\textcite{petkevicPoliticalAttacks2802022} used Krippendorff's $\alpha_K$ and
\textcite{klingerAreCampaignsGetting2023} used the Brennan-Prediger coefficient, which is
less sensitive to class imbalance than Cohen's kappa. We
use the same measures to enable direct comparison with these studies.

In both studies, the authors compute IRR only on a small subset of the
data and in \textcite{klingerAreCampaignsGetting2023} only for the UK subset. We instead
evaluate classification performance on the full datasets.

It is important to note that the Klinger et al. (2023) dataset relies on single-coder annotations at the country level, and that reported intercoder reliability is based on a subset of the UK data. As a result, the benchmark does not provide a fully comparable estimate of human–human agreement across all languages and cases included in our evaluation. We therefore interpret the comparison conservatively: as an assessment of whether LLM classifications align with the available human-coded standard, rather than as a direct comparison between model performance and a fully established human benchmark based on identical double-coded data.

For the LLM annotation, we apply two OpenAI models: gpt-4o-2024-08-06
and gpt-4o-mini-2024-07-18. (The Supplementary Material includes
analysis using OpenAI's more recently released gpt-4.1 models, said to
feature improved instruction-following and context handling. These show
marginal improvements over GPT-4o. However, as the API costs outweighed
these marginal performance gains, we retain GPT-4o-mini for the main
analysis.) These models show high performance and are easily scalable to
large datasets. GPT-4o represented OpenAI's most
advanced general-purpose model at the time of analysis (excluding specialized
reasoning models), while GPT-4o-mini offered a more cost-effective
alternative with a reduced number of parameters. Temperature is set to 0
to ensure deterministic outputs and reproducibility.

\section{Evaluating LLM Performance for Cross-Country Annotation of Negative Campaigning}
\begin{table}[htbp]
\centering
\caption{Classification performance on the Petkevic and Nai (2022) dataset (U.S. Senate elections, 2018)}
\label{tab:petkevicnai}

\begin{tabularx}{\textwidth}{@{}l l *{4}{>{\centering\arraybackslash}X}@{}}
\toprule
\textbf{Method} & \textbf{Model} & \makecell{\textbf{F1}\\(No neg.)} & \makecell{\textbf{F1}\\(Neg.)} & \raisebox{-7pt}{\shortstack{\textbf{Weighted}\\\textbf{F1}}} & \makecell{\textbf{Kripp.}\\$\boldsymbol{\alpha}$} \\
\midrule
Human coders & -- & -- & -- & -- & 0.790 \\
Supervised model & MLP & 0.810 & 0.830 & -- & -- \\

\addlinespace
\multicolumn{6}{l}{\textit{LLM classification (prompting conditions)}} \\

\makecell[l]{Full prompt\\(system + user + codebook)} & 4o & 0.877 & 0.850 & 0.864 & 0.728 \\
\makecell[l]{Full prompt\\(system + user + codebook)} & 4o-mini & 0.919 & 0.912 & 0.916 & 0.831 \\

\makecell[l]{Reduced prompt\\(system + codebook)} & 4o & 0.904 & 0.891 & 0.900 & 0.795 \\
\makecell[l]{Reduced prompt\\(system + codebook)} & 4o-mini & 0.922 & 0.916 & 0.919 & 0.838 \\

\makecell[l]{Minimal prompt\\(codebook only)} & 4o & 0.932 & 0.928 & 0.930 & 0.860 \\
\makecell[l]{Minimal prompt\\(codebook only)} & 4o-mini & 0.929 & 0.926 & 0.927 & 0.855 \\
\bottomrule
\end{tabularx}

\vspace{0.5em}

\begin{minipage}{0.9\linewidth}
\footnotesize
\textit{Notes:} The positive class corresponds to the presence of negative campaigning (i.e., explicit attack or critique toward an opposing party or candidate), while the negative class refers to its absence. F1 scores combine precision and recall into a single performance metric and are reported separately for each class as well as in weighted form (weighted by class frequency). Krippendorff's $\alpha$ reports agreement with the human-coded benchmark. Human coders evaluated 200 randomly sampled tweets, while the supervised baseline (MLP) was trained and tested on a subset of the data (20\%, $N = 234$). LLM performance is evaluated against the full set of singly coded tweets ($N = 1186$). Prompting conditions vary in the amount of contextual information provided to the model, ranging from a minimal codebook-only prompt to a full prompt including system and user context.
\end{minipage}

\end{table}

We begin by focusing on the data from \textcite{petkevicPoliticalAttacks2802022} covering the
2018 U.S. Senate. These scholars employ a broad definition of negative
campaigning, focusing on ``the presence of an explicit attack or
critique toward an opponent.'' Table 1 shows the result of the
comparison of the LLM against this dataset, revealing exceptionally high
performance of both models.

The context-free, zero-shot LLM outperformed both human coders and the
machine learning method employed by \textcite{petkevicPoliticalAttacks2802022}. It achieved
higher IRR than human coders ($\alpha_K$ = .860 vs .790), and showed substantial
performance improvements compared to Petkevic and Nai's
(2022) machine learning method: weighted-F1 increased from .810 to .932
(+.122) for absence detection and from .830 to .928 (+.098) for presence
detection.

For this task, the LLM thus not only offers a faster, cheaper, and more
accessible method, but in fact also achieves higher performance than both
human coders and task-specific supervised methods.

\begin{table}[htbp]
\centering
\caption{Classification performance on the Klinger et al.\ (2023) dataset (European Parliament campaigns, 2014 and 2019)}
\label{tab:klinger1}

\begin{tabularx}{\textwidth}{@{}l l *{4}{>{\centering\arraybackslash}X}@{}}
\toprule
\textbf{Method} & \textbf{Model} & \textbf{Acc.} & \textbf{$F1_W$} & \textbf{$\alpha_K$} & \textbf{$\kappa_{BP}$} \\
\midrule
Human coders & -- & 0.930 & -- & 0.464 & 0.895 \\

\addlinespace
\multicolumn{6}{l}{\textit{LLM classification (prompting conditions)}} \\

\makecell[l]{Minimal prompt\\(original codebook)} & 4o & 0.892 & 0.922 & 0.301 & 0.784 \\
\makecell[l]{Minimal prompt\\(adjusted codebook)} & 4o & 0.947 & 0.956 & 0.435 & 0.893 \\
\makecell[l]{Minimal prompt\\(original codebook)} & 4o-mini & 0.914 & 0.936 & 0.347 & 0.828 \\
\makecell[l]{Minimal prompt\\(adjusted codebook)} & 4o-mini & 0.963 & 0.962 & 0.371 & 0.927 \\
\bottomrule
\end{tabularx}

\vspace{0.5em}

\begin{minipage}{0.9\linewidth}
\footnotesize
\textit{Notes:} Human inter-coder reliability was assessed using 13 coders who each coded the same 150 Facebook posts from the UK. LLM performance was evaluated against a single coder's annotations on 2,043 posts. Of these, 65 posts (3.1\%) were coded as negative campaigning. Metrics include percentage agreement (accuracy), weighted F1 score ($F1_W$), Krippendorff's alpha ($\alpha_K$), and Brennan--Prediger coefficient ($\kappa_{BP}$). All LLM conditions use a minimal prompt (codebook only). The adjusted-codebook condition revises the prompt to better match the stricter operationalization used in \textcite{klingerAreCampaignsGetting2023}.
\end{minipage}

\end{table}

\begin{table}[htbp]
	\centering
	\begin{tabular}{@{}llllll@{}}
		\toprule
		Country & Acc & $F1_W$ & $\kappa_{BP}$ & Supp$_0$ & Supp$_1$\\ \midrule
		AU & 0.936 & 0.925 & 0.871 & 742 & 81 \\
		DE & 0.969 & 0.968 & 0.937 & 1327 & 46 \\
		ES & 0.956 & 0.944 & 0.912 & 2296 & 98  \\
		FR & 0.976 & 0.973 & 0.952 & 1473 & 35  \\
		HR & 0.964 & 0.960 & 0.929 & 1040 & 29 \\
		HU & 0.886 & 0.867 & 0.771 & 1191 & 148 \\
		IE & 0.990 & 0.992 & 0.980 & 506 & 2 \\
		IT & 0.907 & 0.894 & 0.814 & 2009 & 267 \\
		NL & 0.953 & 0.937 & 0.906 & 366 & 19  \\
		PL & 0.917 & 0.901 & 0.834 & 1187 & 128  \\
		SE & 0.966 & 0.958 & 0.932 & 988 & 43  \\
		\bottomrule
	\end{tabular}
	\caption{LLM classification performance (4o-mini, adjusted codebook definition) by country compared to human coders from \textcite{klingerAreCampaignsGetting2023}. Each country was coded by a single human annotator. Metrics include percentage agreement (accuracy), weighted F1 score ($F1_W$), and Brennan-Prediger coefficient ($\kappa_{BP}$). Support columns show the number of negative (Supp$_0$) and positive (Supp$_1$) cases per country. The LLM classifiers vary in whether they use the same codebook definition as the human coders or an adjusted definition.}
    \label{tab:klinger2}
\end{table}

Turning to the \textcite{klingerAreCampaignsGetting2023} dataset, covering the European
Parliament, we can also evaluate the model's performance across
languages and country contexts. \textcite{klingerAreCampaignsGetting2023} used a stricter
and more elaborate definition that distinguishes between negative
tonality and negative campaigning. This resulted in highly imbalanced
classes, with positive cases reduced to only 3.1\% in the UK subset.
It also resulted in a challenging task for the human coders, who
achieved only moderate agreement (Krippendorff's $\alpha_K = .46$).

The prompt was adapted to reflect the stricter definition, also
providing more labelled examples to capture the more elaborate
conceptualization. Table~\ref{tab:klinger1} shows the resulting performance of the model.
As the table reveals, the annotation task is substantially more
challenging. Despite its lower performance in absolute terms, the LLM
still performed comparably to human coders across all languages. (It
should furthermore be noted that the model's performance is limited
by the quality of the manually coded data that is used as gold
standard.)

Centrally, classification performance remains consistent across all languages without requiring language-specific adjustments (see Table \ref{tab:klinger2}). Compared to the data manually coded by native speakers, the model achieved a weighted F1 of at least .90 and Brennan-Prediger
coefficient of at least .80 for all languages except Hungarian (weighted F1 = .89, BP = .76). This is consistent with research showing that Hungarian is particularly challenging for LLMs, possibly because it does not belong to the Indo-European language family \parencite{yang2025openhueval}. The smaller model again performed comparably to the full model,
consistent with the US Senate results.

These results demonstrate that, with appropriate prompting, LLMs can achieve levels of agreement with human coding comparable to those observed in existing manually coded datasets, and outperform conventional machine
learning techniques. The model moreover maintained consistent
performance across all languages when evaluated against native speakers,
suggesting the possibility of using it to produce data for comparative
research. The more elaborate and stricter definition used by \textcite{klingerAreCampaignsGetting2023}
proved more challenging for both human coders and the LLM, but the LLM
still achieves high performance.

These findings have relevant implications for the study of negative
campaigning and, more broadly, for comparative research in political
communication. The ability of LLMs to reliably identify nuanced concepts
in textual data across multiple languages opens new possibilities for
large-scale, cross-national analysis that were previously out of reach
due to methodological and resource constraints. 

\section{Why do parties engage in negative campaigning?}
Building on the now demonstrated accuracy of LLM-based classification, we now apply this approach to a large-scale empirical study of negative campaigning. In doing so, we shift from methodological validation to our substantive contribution: explaining why parties differ in their use of negative campaigning.

Research on the strategic use of negative campaigning has expanded considerably in recent 
years, moving beyond its original U.S.-centric focus toward increasingly systematic 
comparative inquiry \parencite{naiWalter2015,haselmayer2019negative}. Yet this growing 
body of work has relied predominantly on expert surveys and self-reported data, and 
large-scale analyses drawing on direct behavioral measures of campaign communication—
particularly in multilingual, multiparty settings—remain limited.

Scholars broadly agree that political actors engage in negative campaigning based on 
strategic calculations, weighing anticipated gains against potential backlash 
\parencite{lauNegativeCampaigningAnalysis2004a,naiWalter2015}. Attack messages can 
lower public evaluations of rivals and thereby boost one's own standing 
\parencite{pinkletonEffectsNegativeComparative1997}, but they also risk backfiring and 
damaging the sender's credibility \parencite{roeseBacklashEffectsAttack1993}. The balance 
of these risks and rewards, however, is not fixed: various factors shape when negativity 
is more or less attractive as a campaign strategy. Existing research has primarily 
emphasized two categories of influence---individual-level characteristics of candidates 
and contextual features of the electoral environment.

At the micro level, candidates' political profiles, personality traits, perceptions of 
campaign dynamics, social backgrounds, and resource availability influence their 
likelihood of going negative \parencite{maierMappingDriversNegative2023}. At the systemic 
level, features such as electoral disproportionality, party system fragmentation, and 
polarization have been found to affect campaign tone across countries 
\parencite{pappMacroLevelDrivingFactors2019}. Further research links higher levels of 
negativity to majoritarian electoral rules, greater income inequality, ethnic divisions, 
and more individualistic societies \parencite{maierWhenConflictFuels2022}.

Less attention has been paid to how \emph{party-level} characteristics systematically 
shape the use of negative campaigning, particularly using direct behavioral evidence from 
multiparty systems outside the United States. Much of what is known from that context 
stems from studies of American elections, where majoritarian rules and a two-party 
system encourage zero-sum competition. Within this setting, two robust predictors emerge. 
First, trailing candidates are more likely to go negative, while frontrunners often avoid 
risky strategies \parencite{damoreCandidateStrategyDecision2002,lauNegativeCampaigningUS2001,
walterWhenStakesAre2014,maierWhenCandidatesAttack2017}. Second, parties in government 
tend to emphasize achievements and avoid attacks due to higher reputational risks, while 
challengers---often with less visibility and governing experience---rely more on aggressive 
tactics \parencite{druckmanCampaignCommunicationsUS2009,haynesAttackPoliticsPresidential1998,
fridkinVariabilityCitizensReactions2011}.

Comparative work has extended these insights beyond the U.S. context. Studies drawing 
on expert surveys and content analyses show that party characteristics---including 
ideology, governing status, and coalition potential---are often stronger predictors of 
campaign negativity than macro-level electoral context 
\parencite{walterWhenGlovesCome2013,walterWhenStakesAre2014}. More recently, Maier, Nai, 
and Verhaar \parencite*{maierNaiVerhaar2024} compared campaign content across 150 parties 
in 28 countries, finding that ideologically extreme parties consistently employ more 
negative tones, while governing status shapes negativity differently in national versus 
European Parliament elections. Yet these studies rely primarily on expert assessments 
of campaign tone, which---as noted above---can constrain measurement granularity and 
validity \parencite{lindstadtWhenExpertsDisagree2020}. Systematic cross-national analysis 
linking party characteristics to negativity as expressed in parties' own digital 
communication remains an underexplored area.

These dynamics differ substantially in multiparty systems, where parties---rather than 
individual candidates---are the primary actors, and where the path to power often involves 
coalition-building. In proportional representation systems, electoral success does not 
guarantee executive authority. Campaign strategies must therefore account for 
post-election negotiations: overly negative campaigning, especially toward ideologically 
adjacent actors, may damage coalition prospects \parencite{walterExplainingUseAttack2015}. 
Moreover, voters in such systems have more alternatives, which increases the risk of 
defection in response to negativity: negative attacks may destabilize the target while 
benefiting uninvolved third parties rather than the attacker 
\parencite{walterUnintendedConsequencesNegative2019,galassoPositiveSpilloversNegative2023,
mendozaFleetingAllureDark2024}.

Under these conditions, familiar predictors like electoral competitiveness or incumbency 
offer limited explanatory power. Parties with few votes may still enter government 
through coalition deals, and the conventional incumbent--challenger divide becomes less 
relevant. A more useful distinction is coalition potential---the likelihood that a party 
will be included in government \parencite{elmelund-praestekaerAmericanNegativityGeneral2010,
walterWhenGlovesCome2013}. This shifts the incentives: parties with governing prospects 
must balance criticism with the need to remain viable coalition partners, while those 
without such prospects can more freely adopt confrontational styles.

This divide often maps onto the distinction between mainstream and challenger parties 
\parencite{vriesPoliticalEntrepreneursRise2020}. Mainstream parties are more likely to hold 
cabinet office and have a reputational interest in appearing competent and cooperative. 
Challenger parties---including populist radical-right parties, which tend to frame 
politics as an antagonistic contest between a corrupt elite and the people---are less 
integrated into governing coalitions and freer to use aggressive rhetoric 
\parencite{valliNai2020,maierNaiVerhaar2024}. Evidence suggests that ideological extremity 
and opposition status are among the most consistent party-level predictors of negative 
campaigning across comparative settings.

To explain variation in negativity among parties, we propose a framework grounded in 
strategic incentives. Institutional position shapes both the potential costs of going 
negative and the need to cooperate post-election. Governing parties, or those seeking 
future participation in coalitions, are more likely to avoid antagonistic strategies to 
preserve their credibility and relationships \parencite{walterWhenGlovesCome2013}. In 
contrast, outsider or opposition parties face fewer such constraints and can use 
negativity to gain attention and challenge the status quo.

\medskip\noindent\textbf{H1}: Parties in cabinet at the time of communication engage in less negative campaigning.\medskip

Strategic incentives are also shaped by parties' ideological profiles
and communicative styles. Parties at the ideological margins often
define themselves in opposition to the political center and use
confrontational tactics to differentiate themselves. This oppositional
stance is particularly pronounced when their coalition prospects are
slim and reputational risks are minimal \parencite{elmelund-praestekaerAmericanNegativityGeneral2010, walterWhenGlovesCome2013}. Supporting this, \textcite{maierMappingDriversNegative2023}
show that ideologically extreme candidates are more likely to attack,
while \textcite{pappMacroLevelDrivingFactors2019} link higher system polarization to a more
negative tone.

\medskip\noindent\textbf{H2}: Parties further from the ideological center engage in more negative campaigning.\medskip

Populist parties exemplify conflict-driven communication. By framing
politics as a struggle between a virtuous people and a corrupt elite \parencite{muddePopulistZeitgeist2004},
populists are incentivized to adopt an antagonistic
style. Their outsider identity and anti-establishment discourse
predispose them to use attacks more frequently, especially in media and
digital arenas where such messages can rapidly attract attention \parencite{engesserPopulismSocialMedia2017,engesserPopulistOnlineCommunication2017, braccialeDefinePopulistPolitical2017}. Negativity is
thus central to how populist parties can define and perform their
political identity.

\medskip\noindent\textbf{H3}: Populist parties engage in more negative campaigning.\medskip

Among populist actors, radical right populist parties are particularly
incentivized to employ negativity. Their discourse is often driven by
exclusionary nationalism, anti-immigration rhetoric, and a heightened
sense of threat to the cultural or economic status quo---elements that
naturally lend themselves to adversarial messaging. Moreover, their
reliance on mobilizing resentment and moral outrage reinforces a style
of campaigning that is intensely confrontational \parencite{muddePopulistRadicalRight2007, moffittGlobalRisePopulism2016}.

\medskip\noindent\textbf{H4}: Radical right populist parties engage in more negative campaigning than other populist parties.

\subsection{Data and method}\label{data-and-method}

We draw on \textcite{vanvlietTwitterParliamentarianDatabase2020}, which includes 18,066,672 tweets by 5,439 parliamentarians from 2017 to 2022. The dataset covers all tweets from parliamentarians with Twitter accounts in EU member states, candidate states, and EFTA countries where a substantial share of legislators use the platform. We focus on Austria, Belgium, Switzerland, Germany, Denmark, Spain, Finland, France, the United Kingdom, Greece, Ireland, Iceland, Italy, Latvia, the Netherlands, Norway, Poland, Slovenia, and Sweden.

The selection of countries is primarily driven by data availability. To our knowledge, this dataset provides the most comprehensive and systematically comparable source of cross-national data on parliamentary communication currently available. Importantly, it focuses on countries in which Twitter constituted a meaningful arena of elite political communication. This helps ensure that the observed communication patterns are informative of broader party strategies rather than idiosyncratic behavior by a few highly active individuals.

At the same time, the countries included span considerable variation in party systems, electoral institutions, and political dynamics, providing meaningful comparative leverage within a broadly comparable regional context. Focusing on this set of countries also allows us to observe communication across multiple electoral cycles and political periods—including the COVID-19 pandemic—within a consistent data-generating framework.

Twitter (now X) has historically occupied a distinctive position in political communication and research \parencite{jungherrTwitterUseElection2016}. Its affordances for brief, direct messaging that bypassed traditional media gatekeepers contributed to widespread adoption among journalists and politicians, extending its influence well beyond the platform itself \parencite{enliPersonalizedCampaignsPartyCentred2013, oschatzTwitterNewsAnalysis2022}. Unlike many other social media platforms, Twitter also facilitated direct interactions between political actors, including both collaborative exchanges and adversarial debate \parencite{vanvlietTwitterParliamentarianDatabase2020, keucheniusWhyItImportant2021}. These features have made it a particularly valuable source for analysing party positions and communication strategies \parencite{haselmayerNegativeCampaigningIts2019}. At the same time, it is important to note that this position has evolved in recent years. Changes in platform governance, user composition, and patterns of use mean that Twitter no longer occupies the same central role in political communication as it did during the period studied here \parencite{tornberg2025shifts}. Our analysis should therefore be understood as capturing a specific phase in the development of digital political communication, rather than necessarily reflecting the current media environment.

Based on the validation results, we employ the prompt used to replicate
\textcite{petkevicPoliticalAttacks2802022}, as this most closely aligns with the
conventionally used definition of \textcite{geerDefenseNegativityAttack2006}. We make one minor
adjustment, adding `party' to the
definition to better account for the more important role of parties in
multiparty systems. The final prompt hence seeks to identify:
`the presence of explicit attack or critique toward opponent party or candidate.' (See also Supplementary
Material.) Since the validation showed that the smaller model performed
as well as or better than the larger model at a fraction of the cost, we
employ gpt-4o-mini for the full dataset. The classification of the
complete dataset cost US\$156\footnote{Using the gpt-4o-mini-2024-07-18 model with batch API costing \$0.075 per 1M input tokens and \$0.30 per 1M output tokens on the 18,066,672 tweets, requiring 1,938,825,699 input tokens and 36,464,655 output tokens.}.

Due to data availability, only ten of the sixteen languages included in the study were directly validated against human-coded benchmarks. While the results indicate consistent classification performance across the validated languages, extending the analysis to the full dataset requires the assumption that this performance generalizes to the six unvalidated languages. This introduces a degree of methodological uncertainty. However, there is limited reason to expect substantial degradation. Prior research shows that large language models tend to underperform primarily in low-resource or less widely used languages \parencite{rathjeGPTEffectiveTool2024}, a concern that is less likely to apply to the remaining European languages in our sample.

To address this concern, we estimate all models separately for (i) the subset of validated languages and (ii) the full set of languages, allowing us to assess the robustness of the results to potential variation in classification quality. The estimates based on the validated-language subset should be understood as the most conservative benchmark, and we treat these as our preferred reference when interpreting the results.

For the inferential analysis, we employed an OLS regression model to
examine the predictors of negative campaigning, with the percentage of
negative posts per party as the dependent variable. While modeling
proportions with OLS can raise concerns---such as predictions falling
outside the {[}0,1{]} range and potential non-linearity near the
boundaries---these issues are limited in our case, as observed values
range from 0.03 to 0.68. Robustness checks using logit models are
provided in the supplementary material. For improved readability of the results, we multiplied the dependent variable by 100 in the OLS models. To account for party-level
characteristics that may shape negative campaigning, we enriched the
dataset with information from the Chapel Hill Expert Survey \parencite{jollyChapelHillExpert2022a} on ideology and party family, ParlGov \parencite{doringParlGov2024Release2024} on cabinet status and election dates, and the Populism and Political Parties Expert Survey (POPPA) \parencites{zasloveStatePopulismIntroducing2025, meijersMeasuringPopulismPolitical2021} on the degree of populism of parties.

We note that expert surveys are used here for a conceptually different purpose than the one we critique above. Our concern with expert surveys relates to their use as proxies for campaign negativity itself, where they replace direct measurement of textual communication. In contrast, we use expert survey data (e.g., CHES) to capture relatively stable party-level attributes such as ideology, party family, and populism. These variables are not directly observable from text alone and are standardly measured using expert assessments. 

We focus on party-level characteristics derived from our theoretical framework, including ideology, populism, and governing status. We do not include additional controls such as party size (e.g., vote or seat share) or system-level polarization indices. Party size is closely related to governing status and electoral competitiveness, and including it risks introducing post-treatment bias or over-controlling for factors that are themselves shaped by party strategy. Similarly, polarization indices capture system-level features that are partly endogenous to party positioning and competition. Our aim is therefore not to estimate a fully saturated model, but to test theoretically motivated relationships while maintaining interpretability. 
We therefore leave the inclusion of such variables to future research.

The model includes six key predictors:

\begin{enumerate}
\def\labelenumi{\arabic{enumi}.}
\item
  Left--Right Ideology (LRGEN): This variable captures the party's
  overall ideological orientation on a 0--10 scale, where 0 indicates
  the extreme left and 10 the extreme right.
\item
  Ideological Extremism (Non-Centrist Position): Defined as the squared distance from the ideological center, this variable is calculated as $(LRGEN - 5)^2$ and ranges from 0 to 25, with higher values indicating greater ideological extremism regardless of direction. The squared transformation is non-linear: each additional step away from the center (5) carries greater weight than the previous, such that parties at the extremes of the scale are weighted more heavily than those near the center.
\item
  Cabinet party: A binary indicator whether the tweet was posted by a party that at the time of posting was part of the cabinet (1 = cabinet party at time of posting; 0 = opposition).
\item
Election proximity: A time-varying binary indicator of whether a tweet was posted within three months preceding parliamentary elections (1 = election proximity; 0 = non-election proximity). For France, Finland, and Poland, presidential elections are included given their political salience.
\item
  Populism (populism\_mean): A continuous measure (0–10) from the POPPA expert survey, operationalising populism as a latent construct following the ideational approach \parencite{muddePopulistZeitgeist2004}. It is the factor-weighted mean of five expert-rated dimensions: anti-elitism, people-centrism, Manichean worldview, indivisible people, and general will.
\item
  Party Family (FAMILY): Categorical classification of parties into one
  of 11 ideological families. The typology originates from \textcite{hixPoliticalPartiesEuropean1997}
  and has been refined by the CHES team.
\end{enumerate}

The Twitter data were collected between 2017 and 2022, whereas the CHES
data refer to 2019 and 2024, and the POPPA data to 2018 and 2023. This temporal mismatch introduces a degree of
imprecision in the independent variables, as they do not capture
within-period variation or temporal shifts. However, given that the CHES and POPPA data fall near the midpoint of the Twitter collection period, and that the indicators reflect relatively stable party characteristics,
this limitation is considered minor.

Some potentially relevant time-sensitive factors---such as electoral
competitiveness or polling trends---are not included. However, prior
research suggests that such campaign-specific dynamics play a more
limited role in multiparty systems \parencite{walterWhenStakesAre2014, elmelund-praestekaerAmericanNegativityGeneral2010}. Moreover, the focus of this study is on structural, longer-term party characteristics rather than short-term electoral fluctuations.

To account for unobserved heterogeneity at the national level, country
fixed effects are included. Standard errors are clustered at the country
level to address the nested structure of the data. As a robustness
check, a multilevel version of the model was also estimated; results are
reported in the Supplementary Material. Finally, independents and
parties with fewer than 500 tweets were excluded from the analysis to
reduce estimation noise due to high variance in small samples.

While our dataset consists of individual-level messages posted by parliamentarians, we aggregate these data to the party level for analysis. This choice reflects our theoretical focus on party strategy. In parliamentary systems, elected representatives are typically embedded within party organizations and operate under varying degrees of party discipline and coordination. As a result, their public communication is not purely individual but often shaped by shared party goals, strategic incentives, and collective positioning \parencite[e.g.,][]{cox1993legislative,bowler1999party,sieberer2006party}. This is particularly the case on highly visible platforms such as social media, where messages are frequently interpreted by audiences, journalists, and opponents as indicative of the party's stance rather than that of a single politician.

Aggregating to the party level therefore allows us to capture systematic patterns in 
communication that reflect party-level strategy and to relate these patterns directly to 
party characteristics—such as ideology, populism, and governing status—that are central 
to our theoretical framework. This approach is also consistent with a large body of 
comparative research showing that party-level attributes are key predictors of political 
behavior and communication, particularly in proportional and coalition-based systems 
\parencite[e.g.,][]{klingemann2006parties,schumacher2013left}.

To account for temporal effects such as changing institutional position and electoral cycles, we aggregate tweets to the party level separately for each combination of cabinet status (in government vs.\ in opposition) and election proximity (within or outside the three-month pre-election window). This means that if a party changed cabinet status and also had elections during the data collection period, it can appear at most four times in the dataset---once for each of the four possible combinations of these two binary indicators.

Several limitations of this approach merit attention. First, while we account for temporal effects, the analysis abstracts from more fine-grained, context-specific dynamics such as election campaigns, scandals, or political crises. Electoral periods are treated in a harmonized way across countries, which facilitates cross-national comparison but may obscure short-term fluctuations in negativity. Future research could build on this by examining how negative campaigning varies over time and in response to country-specific political events.

Second, our dataset includes all tweets posted by parliamentarians, rather than only explicitly political messages. This inclusive approach provides a comprehensive view of their communication practices, but also incorporates non-political content—such as greetings or personal updates—into the analysis. As a result, our estimates of negativity are likely to be conservative. Studies focusing exclusively on political communication would likely find higher levels of negative campaigning. While capturing the broader communication repertoire of politicians is analytically valuable, caution is therefore warranted when comparing our results to studies based on more narrowly defined campaign messages. Additionally, due to the setup of the data collection, only tweets of national parliamentarians are included, excluding potentially influential political figures without parliamentary representation. Notable examples include \textit{Reform UK} in the United Kingdom and \textit{Gibanje Svoboda} in Slovenia.

A further limitation concerns the scope of platform use. Not all parties and politicians use Twitter to the same extent, and levels of activity vary systematically across countries, party families, and individuals. The dataset therefore reflects the behavior of those actors who are active on the platform, rather than the full universe of political communication. In addition, while Twitter has been a particularly important arena for elite political communication—especially during the period studied—its affordances are specific: communication is primarily text-based, public, and interaction-oriented. The findings should therefore be interpreted as most directly applicable to this type of communication environment, and may not generalize straightforwardly to more visual or multimodal platforms, such as Instagram or TikTok, where political messaging relies more heavily on images, video, and platform-specific formats.

Finally, our measure captures the \textit{presence} of negative campaigning but does not distinguish between different targets or directions of attacks. We leave the analysis of such distinctions for future research.

\subsection{Results}\label{results}

\begin{figure}[htbp]
    \begin{center}
        \caption{Usage of negative campaigning on Twitter between 2017 and 2022 across European countries, split by original tweets and retweets.\\
        \small\textit{Alt text: Dot plot comparing rates of negative campaigning in tweets and retweets across 19 European countries. Slovenia, Spain, and Greece show the highest rates (30--35\%), while Ireland and Iceland show the lowest (around 8--9\%). In most countries, retweets exhibit higher negativity than original tweets.}}
        \label{fig:rt_vs_nort}
        \includegraphics[width=0.8\textwidth]{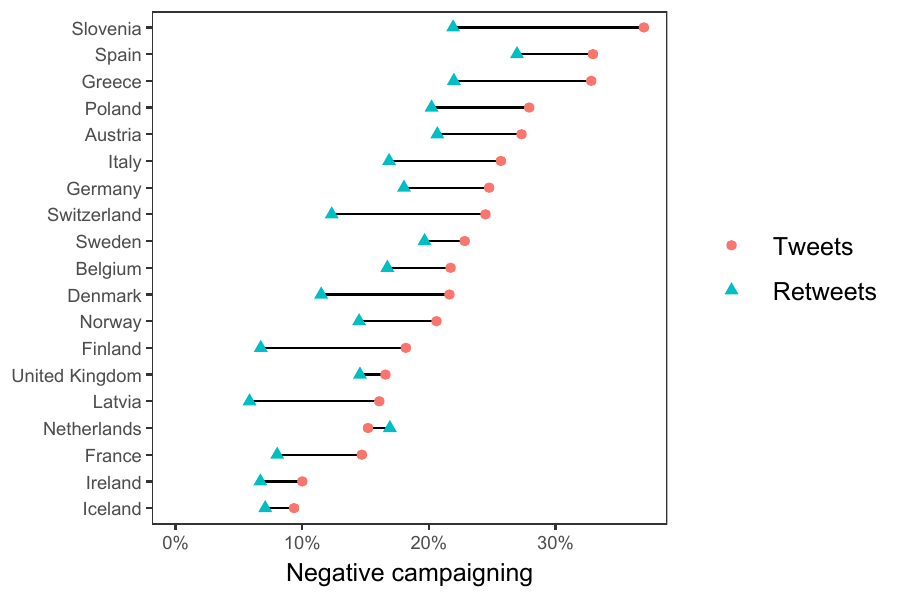}
    \end{center}
\end{figure}

To assess the level of negative campaigning across European states,
Figure~\ref{fig:rt_vs_nort} shows the percentage of negative tweets by country,
distinguishing between original tweets and retweets. As the figure
shows, original tweets tend to be characterized by more negative
campaigning than retweets in all countries except the Netherlands. This corroborates the finding by \textcite{lichtMeasuringUnderstandingParties2024} that retweets
often contain non-political content such as follower spam or event links
rather than political speech. While some politicians, notably Trump in
the US \parencite{brookingsinstitutionHowHateMisinformation2020} and Wilders in the Netherlands \parencite{nosAnderePartijenNiet2024}, have strategically used retweets to amplify extreme viewpoints
from supporters, this pattern appears uncommon across European
parliamentarians in the dataset.

Figure \ref{fig:rt_vs_nort} reveals substantial variation in the levels of negative
messaging in tweets across countries, ranging from lows of 9 to 10
percent for Iceland and Ireland to highs of 32 to 37 percent for Spain
and Slovenia.

These findings reflect two important factors that may lead to varying
levels of negative campaigning. First, national social norms \parencite{oschatzThatsNotAppropriate2024} and political culture \parencite{debusNegativeCampaignStatements2024} create
different expectations for negative campaigning across countries.
Depending on the national context, negative campaigning is more accepted
or less likely to be punished by voters \parencite{oschatzThatsNotAppropriate2024}. The
structure of political coalitions also matters. In ideologically
homogenous coalitions, voters may even reward attacks on coalition
partners more \parencite{debusNegativeCampaignStatements2024}. Second, levels of political
professionalisation vary considerably. Larger countries show much higher
tweet volumes per party, with the UK leading at 2.4M tweets, followed by
France (0.9M), Poland (0.7M), and Germany (0.6M) (see also Supplementary
Material). More professionalised campaigns with dedicated social media
teams may deploy negative campaigning more strategically. (Following
\textcite{lichtMeasuringUnderstandingParties2024}, we focus only on original tweets in subsequent
analyses.)

\begin{figure}[htbp]
    \begin{center}
        \caption{Country-level mean share of tweets containing negative campaigning by parties across 19 European democracies on Twitter (2017–2022), disaggregated by message type. Markers indicate country means for original tweets and retweets, respectively; horizontal segments connect the two conditions within each country.\\
        \small\textit{Alt text: A grid of 19 dot plots, one per country, showing the share of negative campaigning tweets by political party. Each panel lists parties on the y-axis and negative campaigning rate (0--60\%) on the x-axis. Across countries, far-right and populist parties (e.g.\ AfD in Germany, FPO in Austria, Vox in Spain, EL in Greece) tend to cluster at the higher end, while green and social-democratic parties generally show lower rates. Substantial within-country variation is visible in most panels.}}
        \noindent\makebox[\textwidth]{%
            \includegraphics[width=1.2\textwidth]{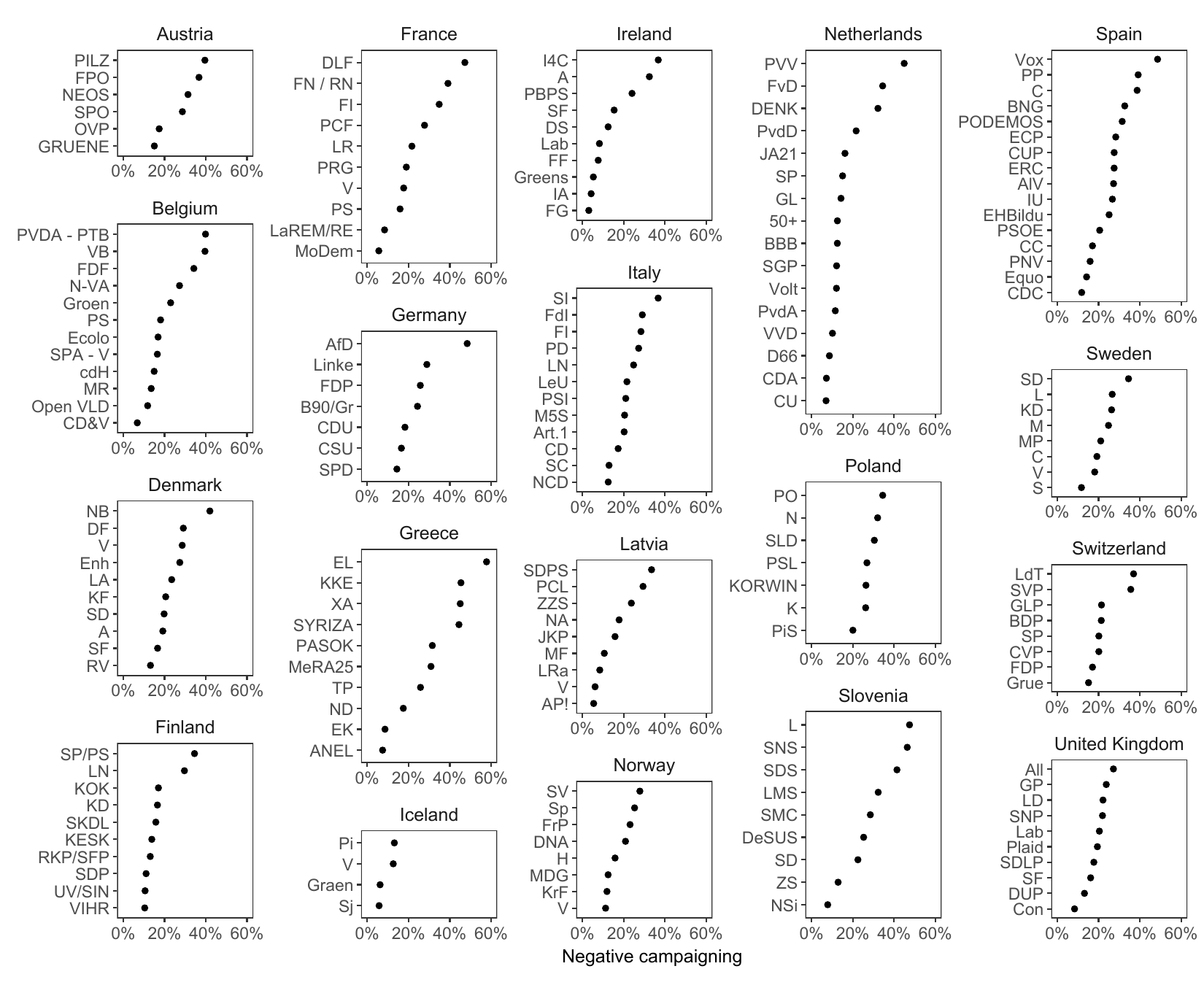}}
        \label{fig:party_country}
    \end{center}
\end{figure}

To further investigate the relationship between parties and negative
campaigning, Figure~\ref{fig:party_country} shows average negative campaigning per party
within countries.

The figure reveals several patterns. First, governing parties or those
with government experience generally show lower negative campaigning
levels. This is evident in the Netherlands where the four coalition
parties (VVD, D66, CDA, CU) have the lowest rates, as well as in Germany
(CDU, CSU, SPD), Poland (PiS), and Sweden (S). Meanwhile, challenger or
opposition parties tend to have the highest levels within countries.

Second, there is no strong association between political ideology and
the likelihood of engaging in negative campaigning, although right-wing
parties tend to use such strategies somewhat more frequently. For
instance, radical right parties exhibit the highest probabilities of
negativity in Denmark (\emph{Nye Borgerlige}) and Sweden
(\emph{Sverigedemokraterna}), while in Norway, the left-wing
\emph{Sosialistisk Venstreparti} stands out as the most negative. In
Greece, the right-wing \emph{Elliniki Lisi} (EL) leads, whereas in
Belgium, the left-wing \emph{PVDA-PTB} ranks highest, followed closely
by the right-wing \emph{Vlaams Belang}. In Germany, the radical right
\emph{Alternative für Deutschland} shows the highest level of
negativity, but it is closely followed by the left-wing \emph{Die
Linke.}

Additionally, these results hint at deliberate strategic use of negative
campaigning. In Switzerland, \emph{Lega dei Ticinesi} and
\emph{Schweizerische Volkspartei}, parties with electoral alliances,
show similarly high levels of negative campaigning. Similarly,
Germany's \emph{CDU} and \emph{CSU} demonstrate
coordinated low negativity.

The UK stands out for its noticeably lower levels of negative
campaigning, despite being a majoritarian system that should
theoretically produce higher negativity. Two factors may explain this:
first, UK politicians are frequent users of Twitter (shown in the high
tweet volume, with 2,021,218 messages) which may imply that many
messages are constituency-focused rather than partisan, thus diluting
the percentage of negative tweets. Second, challenger parties like
\emph{UKIP}, \emph{Brexit Party} and \emph{Reform UK}, likely to have higher
negativity rates, are missing because they were not in national
parliament during data collection.

\begin{figure}[htbp]
  \centering
  \begin{subfigure}[t]{0.48\textwidth}
    \centering
    \includegraphics[width=\textwidth]{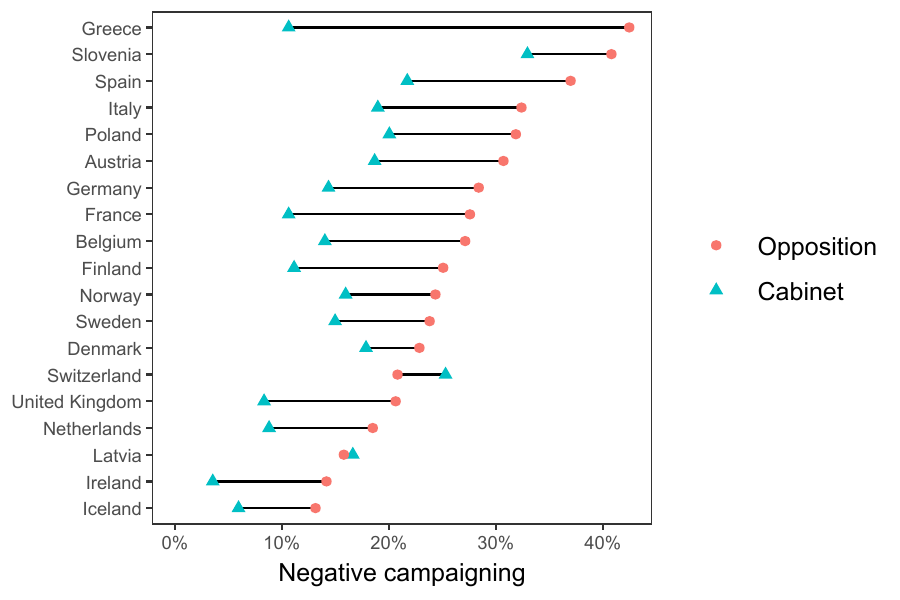}
    \caption{Cabinet vs. Opposition.}
    \label{fig:cabinet_vs_no_cabinet}
  \end{subfigure}
  \begin{subfigure}[t]{0.48\textwidth}
    \centering
    \includegraphics[width=\textwidth]{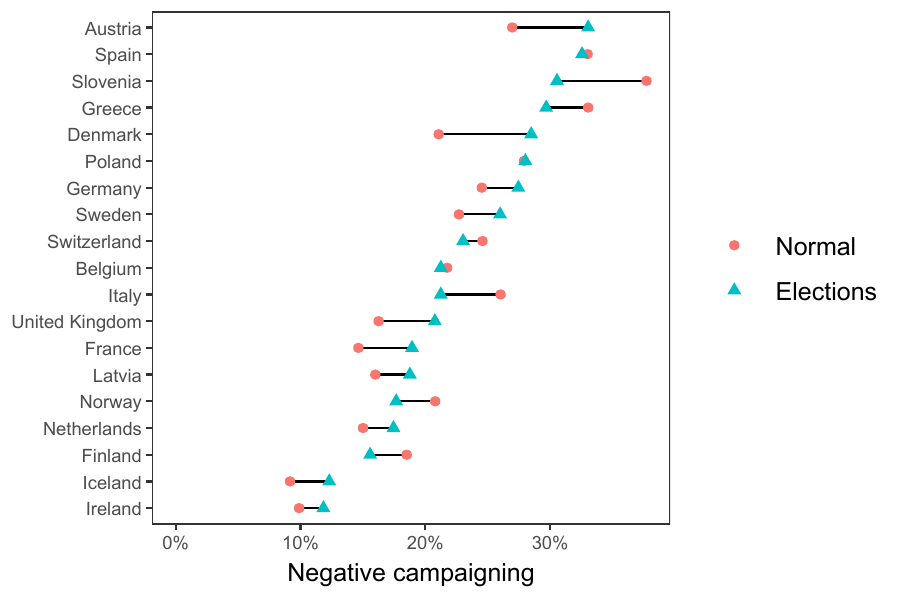}
    \caption{Election period vs. no election.}
    \label{fig:election_vs_no_election}
  \end{subfigure}
  \caption{Country-level mean share of tweets containing negative campaigning, by (a) cabinet status at the time of posting and (b) proximity to elections (tweets posted within three months of election day). Markers indicate country means, and horizontal segments connect the two conditions within each country. In panel (a), circles denote opposition parties and triangles cabinet parties; in panel (b), circles denote non-election periods and triangles election periods. The figure is descriptive and based on raw averages; adjusted estimates are reported in the multivariate models below.\\
  \small\textit{Alt text: Two side-by-side dot plots comparing negative campaigning rates across 19 European countries. Panel (a) shows opposition parties (circles) nearly always scoring higher than cabinet parties (triangles), with Greece, Slovenia, and Spain at the top. Panel (b) shows election periods (triangles) associated with slightly higher negativity than normal periods (circles) in most countries, though the gap is smaller and less consistent than in panel (a).}}
  \label{fig:negative_campaigning}
\end{figure}

Figure \ref{fig:negative_campaigning} provides descriptive evidence on the two time-varying political conditions introduced in the main models: cabinet status and election proximity. Panel (a) compares the average share of negative tweets posted by parties in cabinet and outside cabinet at the time of posting. In nearly all countries, cabinet parties exhibit lower levels of negative campaigning than non-cabinet parties. This pattern offers descriptive support for H1 and is consistent with the argument that governing parties face greater reputational and coalition-management costs when adopting antagonistic rhetoric. Switzerland is the main exception, with cabinet parties appearing somewhat more negative than non-cabinet parties.

Panel (b) compares the average share of negative tweets posted during election periods---defined as the three months prior to an election---with tweets posted outside such periods. Here the pattern is less uniform. In many countries, negativity is somewhat higher during election periods, but the size of the difference varies substantially across contexts, and in a few cases the pattern reverses. This suggests that election proximity may intensify negative campaigning, but in a more contingent and context-dependent way than the cabinet--opposition distinction. The figure should therefore be read as a descriptive complement to the multivariate analysis below, which estimates these relationships while accounting for other party characteristics.

\begin{table}
\begin{center}
\begin{tabular}{l c c c c}
\toprule
 & Validated 1 & Validated 2 & All 1 & All 2 \\
\midrule
(Intercept)         & $18.06^{*}$        & $21.98^{*}$       & $17.75^{*}$        & $21.35^{*}$        \\
                    & $ [  7.63; 28.49]$ & $ [14.09; 29.86]$ & $ [ 10.54; 24.97]$ & $ [ 15.37; 27.34]$ \\
Cabinet Party       & $-8.28^{*}$        & $-6.86^{*}$       & $-8.59^{*}$        & $-7.56^{*}$        \\
                    & $ [-11.33; -5.24]$ & $ [-9.96; -3.76]$ & $ [-11.21; -5.97]$ & $ [-10.42; -4.70]$ \\
Populism            & $1.97^{*}$         & $1.07$            & $2.08^{*}$         & $1.35$             \\
                    & $ [  0.54;  3.40]$ & $ [-1.13;  3.27]$ & $ [  1.06;  3.10]$ & $ [ -0.20;  2.90]$ \\
Election Proximity  & $1.26$             & $1.26$            & $1.37^{*}$         & $1.29^{*}$         \\
                    & $ [ -0.16;  2.68]$ & $ [-0.21;  2.73]$ & $ [  0.13;  2.61]$ & $ [  0.09;  2.49]$ \\
General Left-Right  & $0.60$             &                   & $0.57^{*}$         &                    \\
                    & $ [ -0.21;  1.41]$ &                   & $ [  0.01;  1.13]$ &                    \\
Ideological Extreme &                    & $0.57$            &                    & $0.48^{*}$         \\
                    &                    & $ [-0.10;  1.23]$ &                    & $ [  0.04;  0.93]$ \\
\midrule
R$^2$               & $0.54$             & $0.57$            & $0.56$             & $0.58$             \\
Adj. R$^2$          & $0.50$             & $0.54$            & $0.52$             & $0.55$             \\
Num. obs.           & $198$              & $198$             & $290$              & $290$              \\
RMSE                & $7.70$             & $7.39$            & $7.62$             & $7.42$             \\
N Clusters          & $12$               & $12$              & $19$               & $19$               \\
\bottomrule
\multicolumn{5}{l}{\scriptsize{$^*$ Null hypothesis value outside the confidence interval.}}
\end{tabular}
\caption{Main Model of the Effect of Party Characteristics on Negative
			Campaigning in Europe, showing for validated languages only (Austria, Belgium, Switzerland, Germany, Spain, France, United Kingdom, Ireland, Italy, The Netherlands, Poland and Sweden) and for all languages (validated languages plus Denmark, Finland, Greece, Iceland, Latvia, Norway and Slovenia) in data collection. Country-fixed effects not displayed. Standard errors are clustered by country.}
\label{table:coefficients}
\end{center}
\end{table}

We now turn to the inferential results, presented in Table \ref{table:coefficients}. Because the validated-language subset provides the most conservative test of our argument, we treat effects that are statistically robust there as our strongest findings, while effects that emerge only in the full sample are interpreted as suggestive rather than definitive. With regard to party-level characteristics, there is strong evidence that government participation is associated with lower levels of negative campaigning. On average, parties holding cabinet office engage in negative campaigning approximately 7--8 percentage points less than opposition parties. Consistent with previous research, governing parties appear more risk-averse, tending to emphasize their governing record rather than attack opponents, as they face greater reputational and coalition-related costs from negative campaigning. Challenger parties, by contrast, are less constrained and more inclined to adopt negative rhetoric. These findings support hypothesis H1.

The analysis provides evidence consistent with hypothesis H2: parties positioned further from the ideological center tend to engage in more negative campaigning. This relationship reaches statistical significance in the full-sample models but falls just short of conventional thresholds in the validated-language models (coefficient = 0.57, 95\% CI: $-$0.10 to 1.23), indicating that some caution is warranted. Notwithstanding this uncertainty, the direction of the association is consistent across all specifications, and the non-centrist positioning effect is likely not symmetric across the spectrum. When ideology is modeled using the left--right scale, parties on the right appear somewhat more inclined toward negative campaigning, though this effect is also not robust in the validated-language models. Taken together, the results point to a pattern in which negativity is lowest among centrist parties and increases toward the extremes, with the more pronounced rise on the right. This pattern is illustrated in Figure~S5 in the Supplementary Information. The party-family analysis further suggests that the asymmetry is primarily driven by radical right parties, while the radical left is also more negative than centrist parties but less sharply so.

When populism is entered alongside the left--right scale (Validated 1), populist parties exhibit higher levels of negative campaigning---approximately 2 percentage points more than non-populist parties---consistent with hypothesis H3. This aligns with expectations that populist actors employ more confrontational and anti-elite rhetoric. However, this effect is no longer statistically significant once ideological extremism is substituted into the model (Validated 2), suggesting that the higher levels of negativity among populist parties are largely accounted for by their non-centrist ideological positioning rather than populism per se. This pattern indicates that populism and ideological extremity capture overlapping dimensions of party positioning, with much of the association between populism and negativity operating through parties' distance from the ideological center.

\begin{figure}[htbp]
    \begin{center}
        \caption{Average predicted level of negative campaigning by party family, controlling for cabinet status, election proximity, and populism. Error bars show 95\% confidence intervals. The confessional party family has only 3 observations, with 2 located in the Netherlands and 1 in Finland. This geographic concentration affects the reliability of standard error estimates for this category, as Finnish is not among the validated languages.\\
        \small\textit{Alt text: Dot plot with confidence intervals showing predicted negative campaigning rates across 11 party families, comparing all languages (circles) and validated languages only (triangles). Radical Right stands out with the highest rate (around 36--37\%), followed by Radical Left (around 28\%). Confessional parties show the lowest rate (around 10--13\%), though with a note of unreliable estimates. Most other party families cluster between 15--25\%. The two series track each other closely across families.}}
        \noindent\makebox[\textwidth]{%
            \includegraphics[width=1\textwidth]{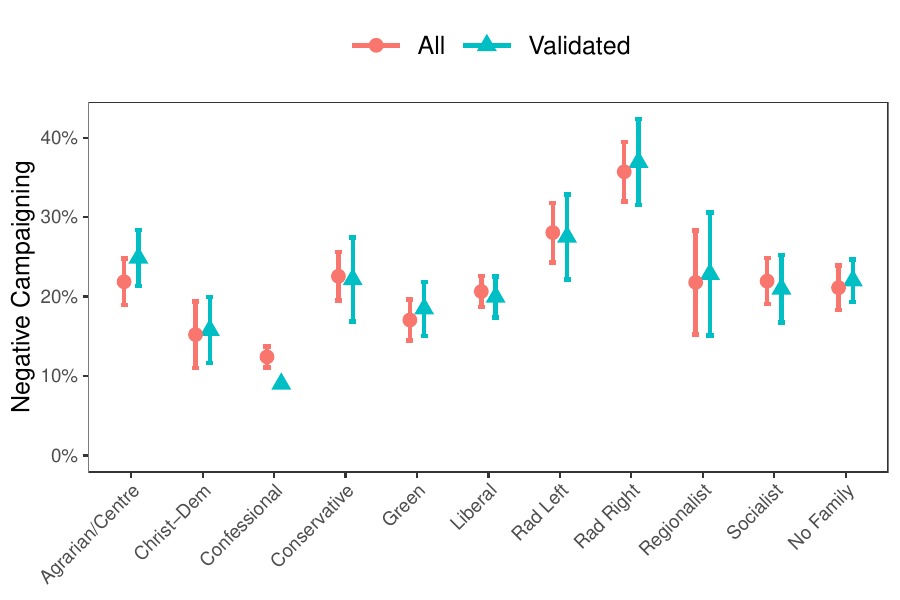}}
        \label{fig:margmeans}
    \end{center}
\end{figure}

To further explore the role of ideology, we examine the effect of party family while controlling for government participation and populist orientation. The continuous ideology variables are excluded from this model to avoid multicollinearity. Figure \ref{fig:margmeans} presents the average predicted levels of negative campaigning across party families, holding cabinet status, election proximity, and populism constant.

The results highlight the distinct position of radical right parties, which display the highest predicted levels of negative campaigning in both samples (36.9\% in the validated-language sample and 35.7\% in the full sample). Radical left parties rank second (27.5\% and 28.1\%, respectively), while Agrarian/Centre (24.9\% and 21.9\%), regional (22.8\% and 21.8\%), and conservative parties (22.2\% and 22.6\%) also exhibit somewhat elevated levels. This pattern is consistent with the expectation that parties farther from the ideological center engage more frequently in negative campaigning, likely because they face weaker coalition incentives and greater strategic benefits from adversarial rhetoric. The radical right, however, is clearly more negative than any other party family, indicating that its communication style is especially confrontational. This supports H4.

\section{Discussion and Conclusion}\label{discussion-and-conclusion}
While negative campaigning has become a prominent topic in political communication research, the field has long faced methodological constraints. Traditional approaches—such as manual annotation or supervised machine learning—are either resource-intensive or limited in their ability to generalize across languages and contexts. As a result, much of the existing literature has focused on single-country settings, particularly the United States, constraining our understanding of how negativity operates across different institutional and cultural environments.

This paper contributes to addressing these limitations in two ways. First, it examines the use of zero-shot LLMs as a tool for identifying negative campaigning. By leveraging their multilingual and contextual capabilities, we show that LLMs can support accurate annotation across languages at relatively low cost, without requiring language-specific training data or extensive manual coding. Second, we apply this approach to conduct a large-scale cross-country study of negative campaigning, providing new comparative evidence on how party-level characteristics shape campaign negativity in multiparty systems.

To assess the reliability of this approach, we evaluate the performance of zero-shot LLMs against established manually coded datasets from \textcite{petkevicPoliticalAttacks2802022} and \textcite{klingerAreCampaignsGetting2023}. The results indicate that LLM-based annotations closely align with human coding and, in our setting, perform comparably to or better than conventional supervised models across multiple languages. These findings are consistent with emerging work \parencite[e.g.,][]{tornbergLargeLanguageModels2024} suggesting that general-purpose language models can perform well on tasks that require contextual interpretation across domains. At the same time, we emphasize that performance depends on careful prompt design, validation against human-coded benchmarks, and transparent reporting of implementation choices.

Beyond performance, the use of LLMs has implications for how text-as-data research is conducted. Manual coding often relies on shared but partly implicit understandings among coders, which can be difficult to fully document in codebooks and may lead to variation across studies. In contrast, LLM-based approaches require researchers to formalize their definitions through prompts, making key analytical decisions more explicit. While this does not resolve underlying conceptual disagreements about how phenomena such as negative campaigning should be defined, it shifts where these disagreements are addressed—by requiring researchers to specify and justify their operationalizations upfront. When prompts, models, and procedures are clearly documented, LLMs can thus contribute to improving transparency and replicability across settings. At the same time, LLM-based annotation introduces its own challenges, including sensitivity to prompt design, potential biases in model outputs, and limited transparency in training data. As such, LLMs are best understood as improving one component of the measurement problem, rather than replacing the need for theoretical clarity and conceptual agreement.

Substantively, the paper contributes to the study of negative campaigning by focusing on party-level dynamics in multiparty systems. While much prior work has emphasized candidate-level or system-level factors, our findings highlight the importance of party characteristics—such as ideology, populism, and governing status—in shaping campaign negativity in proportional and coalition-based contexts.

Using a dataset of 18 million tweets from parliamentarians in 19 countries between 2017 and 2022, we test a framework grounded in strategic incentives. The results suggest that parties located further from the ideological center are more likely to engage in negative campaigning---an association that is consistent across specifications, though most robust in the full-sample models---with the increase particularly pronounced among parties on the radical right. Parties holding cabinet office at the time of communication, in contrast, are less likely to use negative rhetoric, consistent with arguments emphasizing coalition considerations and reputational constraints. These patterns underscore the role of party positioning and institutional context in shaping campaign strategies.

Taken together, the findings demonstrate that LLM-based approaches can be usefully applied to large-scale, cross-country analyses of political communication, while also highlighting the importance of careful validation and cautious interpretation. By lowering some of the practical barriers associated with multilingual text analysis, such methods make it more feasible to study political discourse comparatively and at scale.

At the same time, several limitations should be noted. First, our analysis focuses on textual content and does not capture visual or multimodal forms of campaigning, which are increasingly central to political communication, especially on social media platforms. Second, while we validate the LLM-based annotations against established human-coded datasets, such validation necessarily relies on existing coding schemes and cannot fully eliminate concerns about measurement error or conceptual ambiguity. Third, although the datasets used for validation are not publicly shared, and thus direct inclusion in model training data is unlikely, the opacity of LLM training corpora means that some degree of overlap cannot be entirely ruled out. Fourth, our measure captures the presence of negative campaigning but does not distinguish between different targets or directions of attacks, which may vary systematically across parties and contexts. In addition, the validation relies on existing annotated datasets rather than newly double-coded data, and therefore does not allow a direct comparison between model performance and human–human agreement on identical items across all languages. Finally, while LLMs enable scalable and cross-lingual analysis, their outputs remain sensitive to prompt design and model choice, and future work should continue to assess the robustness of findings across alternative specifications and models. Taken together, these considerations suggest that LLM-based approaches should be seen as a complement to, rather than a replacement for, existing methods, and that continued validation and methodological refinement remain essential.

Future research can extend this approach in several directions, including examining temporal dynamics in negativity, exploring non-democratic contexts, and incorporating visual or multimodal political communication. More broadly, the use of LLMs opens up exciting new possibilities for comparative research, while calling for continued attention to issues of measurement validity, robustness, and transparency.

\section*{Data Availability Statement}
No new data were generated in this study. The data used are available from
\textcite{vanvlietTwitterParliamentarianDatabase2020, klingerAreCampaignsGetting2023},
and \textcite{petkevicPoliticalAttacks2802022}.

\section*{Conflict of Interest Statement}
The authors declare no competing interests.

\section*{Funding}
Funding information has been removed for blind review.

\printbibliography

@book{cox1993legislative,
  title={Legislative Leviathan: Party Government in the House},
  author={Cox, Gary W. and McCubbins, Mathew D.},
  year={1993},
  publisher={University of California Press},
  address={Berkeley}
}

@article{sieberer2006party,
  title={Party unity in parliamentary democracies: A comparative analysis},
  author={Sieberer, Ulrich},
  journal={The Journal of Legislative Studies},
  volume={12},
  number={2},
  pages={150--178},
  year={2006}
}

@book{bowler1999party,
  title={Party Discipline and Parliamentary Government},
  editor={Bowler, Shaun and Farrell, David M. and Katz, Richard S.},
  year={1999},
  publisher={Ohio State University Press},
  address={Columbus}
}

@book{klingemann2006parties,
  title={Parties, Policies and Democracy},
  author={Klingemann, Hans-Dieter and Volkens, Andrea and Bara, Judith and Budge, Ian and McDonald, Michael},
  year={2006},
  publisher={Oxford University Press},
  address={Oxford}
}

@book{naiWalter2015,
  title     = {New Perspectives on Negative Campaigning: Why Attack Politics Matters},
  editor    = {Nai, Alessandro and Walter, Annemarie S.},
  year      = {2015},
  publisher = {ECPR Press},
  address   = {Colchester}
}

@article{haselmayer2019negative,
  title   = {Negative campaigning and its consequences: a review and a look ahead},
  author  = {Haselmayer, Martin},
  journal = {French Politics},
  volume  = {17},
  number  = {3},
  pages   = {355--372},
  year    = {2019},
  doi     = {10.1057/s41253-019-00084-8}
}

@article{valliNai2020,
  title={Attack politics from Albania to Zimbabwe: A large-scale comparative study on the drivers of negative campaigning},
  author={Valli, Chiara and Nai, Alessandro},
  journal={International Political Science Review},
  volume={43},
  number={5},
  pages={680--696},
  year={2022},
  publisher={Sage Publications Sage UK: London, England}
}

@article{maierNaiVerhaar2024,
  title   = {More negative when it matters less? {Comparing} party campaign 
             behaviour in {European} and national elections},
  author  = {Maier, J{\"u}rgen and Nai, Alessandro and Verhaar, Nynke},
  journal = {Journal of European Public Policy},
  volume  = {32},
  number  = {9},
  pages   = {2307--2329},
  year    = {2024},
  doi     = {10.1080/13501763.2024.2404154}
}

@article{galassoPositiveSpilloversNegative2023,
  title   = {Positive Spillovers from Negative Campaigning},
  author  = {Galasso, Vincenzo and Nannicini, Tommaso and Nunnari, Salvatore},
  journal = {American Journal of Political Science},
  volume  = {67},
  number  = {1},
  pages   = {5--21},
  year    = {2023},
  doi     = {10.1111/ajps.12610}
}

@book{vriesPoliticalEntrepreneursRise2020,
  title     = {Political Entrepreneurs: The Rise of Challenger Parties in {Europe}},
  author    = {de Vries, Catherine E. and Hobolt, Sara B.},
  year      = {2020},
  publisher = {Princeton University Press},
  address   = {Princeton, NJ}
}

@article{tornberg2025shifts,
  title={Shifts in US Social Media Use, 2020-2024: Decline, Fragmentation, and Enduring Polarization},
  author={T{\"o}rnberg, Petter},
  journal={arXiv preprint arXiv:2510.25417},
  year={2025}
}

@article{schumacher2013left,
  title={Left-right positions and policy attention: How parties respond to shifts in public opinion},
  author={Schumacher, Gijs and de Vries, Catherine E. and Vis, Barbara},
  journal={Electoral Studies},
  volume={32},
  number={2},
  pages={300--312},
  year={2013}
}

@article{auterNegativeCampaigningSocial2016,
  title = {Negative {{Campaigning}} in the {{Social Media Age}}: {{Attack Advertising}} on {{Facebook}}},
  shorttitle = {Negative {{Campaigning}} in the {{Social Media Age}}},
  author = {Auter, Zachary J. and Fine, Jeffrey A.},
  date = {2016-12-01},
  journaltitle = {Political Behavior},
  shortjournal = {Polit Behav},
  volume = {38},
  number = {4},
  pages = {999--1020},
  issn = {1573-6687},
  doi = {10.1007/s11109-016-9346-8},
  langid = {english}
}

@article{bailCanGenerativeAI2024,
  title = {Can {{Generative AI}} Improve Social Science?},
  author = {Bail, Christopher A.},
  date = {2024-05-21},
  journaltitle = {Proceedings of the National Academy of Sciences},
  volume = {121},
  number = {21},
  pages = {e2314021121},
  publisher = {Proceedings of the National Academy of Sciences},
  doi = {10.1073/pnas.2314021121}
}

@misc{tornberg2026largelanguagemodelsreproduce,
      title={Large Language Models Reproduce Racial Stereotypes When Used for Text Annotation}, 
      author={Petter Törnberg},
      year={2026},
      eprint={2603.13891},
      archivePrefix={arXiv},
      primaryClass={cs.CL},
      url={https://arxiv.org/abs/2603.13891}, 
}

@incollection{baranowski2024patterns,
  title={Patterns of Negative Campaigning during the 2019 European Election: Political Parties' Facebook Posts and Users' Sharing Behaviour across Twelve Countries},
  author={Baranowski, Pawe{\l} and Kruschinski, Simon and Russmann, Uta and Ha{\ss}ler, J{\"o}rg and Magin, Melanie and Bene, M{\'a}rton and Ceron, Andrea and Jackson, Daniel and Lilleker, Darren},
  booktitle={Citizens, Participation and Media in Central and Eastern European Nations},
  pages={26--42},
  year={2024},
  publisher={Routledge}
}

@inproceedings{yang2025openhueval,
  title={OpenHuEval: Evaluating Large Language Model on Hungarian Specifics},
  author={Yang, Haote and Wei, Xingjian and Wu, Jiang and Ligeti-Nagy, No{\'e}mi and Sun, Jiaxing and Wang, Yinfan and Yang, Gy{\H{o}}z{\H{o}} Zijian and Gao, Junyuan and Wang, Jingchao and Jiang, Bowen and others},
  booktitle={Findings of the Association for Computational Linguistics: ACL 2025},
  pages={7464--7520},
  year={2025}
}

@article{weberMeasuringSentimentPolitical2018,
  author    = {Weber, Ren{\'e} and Mangus, J. Michael and Huskey, Richard and Hopp, Frederic R. and Amir, Ori and Swanson, Reid and Gordon, Andrew and Khooshabeh, Peter and Hahn, Lindsay and Tamborini, Ron},
  title     = {Extracting Latent Moral Information from Text Narratives: Relevance, Challenges, and Solutions},
  journal   = {Communication Methods and Measures},
  volume    = {12},
  number    = {2-3},
  pages     = {119--139},
  year      = {2018},
  doi       = {10.1080/19312458.2018.1447656}
}

@article{benoitCrowdsourcedTextAnalysis2016,
  title = {Crowd-Sourced {{Text Analysis}}: {{Reproducible}} and {{Agile Production}} of {{Political Data}}},
  shorttitle = {Crowd-Sourced {{Text Analysis}}},
  author = {Benoit, Kenneth and Conway, Drew and Lauderdale, Benjamin E. and Laver, Michael and Mikhaylov, Slava},
  date = {2016-05},
  journaltitle = {American Political Science Review},
  volume = {110},
  number = {2},
  pages = {278--295},
  issn = {0003-0554, 1537-5943},
  doi = {10.1017/S0003055416000058},
  langid = {english}
}

@article{braccialeDefinePopulistPolitical2017,
  title = {Define the Populist Political Communication Style: The Case of {{Italian}} Political Leaders on {{Twitter}}},
  shorttitle = {Define the Populist Political Communication Style},
  author = {Bracciale, Roberta and Martella, Antonio},
  date = {2017-09-02},
  journaltitle = {Information, Communication \& Society},
  volume = {20},
  number = {9},
  pages = {1310--1329},
  publisher = {Routledge},
  issn = {1369-118X},
  doi = {10.1080/1369118X.2017.1328522}
}

@book{brenneOstrakaVomKerameikos2020,
  title = {Die Ostraka vom Kerameikos},
  author = {Brenne, Stefan},
  date = {2020-12-18},
  journaltitle = {iDAI.publications/books},
  publisher = {iDAI.publications/books},
  doi = {10.34780/kerameikos.v20i0.1000},
  langid = {ngerman}
}

@article{brookingsinstitutionHowHateMisinformation2020,
  entrysubtype = {newspaper},
  title = {How Hate and Misinformation Go Viral: {{A}} Case Study of a {{Trump}} Retweet},
  shorttitle = {How Hate and Misinformation Go Viral},
  author = {{Brookings Institution}},
  date = {2020-09-02},
  langid = {american}
}

@online{brownLanguageModelsAre2020,
  title = {Language {{Models}} Are {{Few-Shot Learners}}},
  author = {Brown, Tom B. and Mann, Benjamin and Ryder, Nick and Subbiah, Melanie and Kaplan, Jared and Dhariwal, Prafulla and Neelakantan, Arvind and Shyam, Pranav and Sastry, Girish and Askell, Amanda and Agarwal, Sandhini and Herbert-Voss, Ariel and Krueger, Gretchen and Henighan, Tom and Child, Rewon and Ramesh, Aditya and Ziegler, Daniel M. and Wu, Jeffrey and Winter, Clemens and Hesse, Christopher and Chen, Mark and Sigler, Eric and Litwin, Mateusz and Gray, Scott and Chess, Benjamin and Clark, Jack and Berner, Christopher and McCandlish, Sam and Radford, Alec and Sutskever, Ilya and Amodei, Dario},
  date = {2020-07-22},
  eprint = {2005.14165},
  eprinttype = {arXiv},
  eprintclass = {cs},
  doi = {10.48550/arXiv.2005.14165},
  pubstate = {prepublished}
}

@article{damoreCandidateStrategyDecision2002,
  title = {Candidate {{Strategy}} and the {{Decision}} to {{Go Negative}}},
  author = {Damore, David F.},
  date = {2002},
  journaltitle = {Political Research Quarterly},
  volume = {55},
  number = {3},
  eprint = {3088036},
  eprinttype = {jstor},
  pages = {669--685},
  publisher = {[University of Utah, Sage Publications, Inc.]},
  issn = {1065-9129},
  doi = {10.2307/3088036}
}

@article{debusNegativeCampaignStatements2024,
  title = {Negative Campaign Statements, Coalition Heterogeneity, and the Support for Government Parties},
  author = {Debus, Marc and Tuttnauer, Or},
  date = {2024-02-01},
  journaltitle = {Electoral Studies},
  shortjournal = {Electoral Studies},
  volume = {87},
  pages = {102738},
  issn = {0261-3794},
  doi = {10.1016/j.electstud.2023.102738}
}

@dataset{doringParlGov2024Release2024,
  title = {{{ParlGov}} 2024 {{Release}}},
  author = {Döring, Holger and Manow, Philip},
  date = {2024-08-12},
  publisher = {Harvard Dataverse},
  doi = {10.7910/DVN/2VZ5ZC},
  langid = {english}
}

@article{druckmanCampaignCommunicationsUS2009,
  title = {Campaign {{Communications}} in {{U}}.{{S}}. {{Congressional Elections}}},
  author = {Druckman, James N. and Kifer, Martin J. and Parkin, Michael},
  date = {2009-08},
  journaltitle = {American Political Science Review},
  volume = {103},
  number = {3},
  pages = {343--366},
  issn = {1537-5943, 0003-0554},
  doi = {10.1017/S0003055409990037},
  langid = {english}
}

@article{elmelund-praestekaerAmericanNegativityGeneral2010,
  title = {Beyond {{American}} Negativity: Toward a General Understanding of the Determinants of Negative Campaigning},
  shorttitle = {Beyond {{American}} Negativity},
  author = {Elmelund-Præstekær, Christian},
  date = {2010-03},
  journaltitle = {European Political Science Review},
  volume = {2},
  number = {1},
  pages = {137--156},
  issn = {1755-7747, 1755-7739},
  doi = {10.1017/S1755773909990269},
  langid = {english}
}

@article{engesserPopulismSocialMedia2017,
  title = {Populism and Social Media: How Politicians Spread a Fragmented Ideology},
  shorttitle = {Populism and Social Media},
  author = {Engesser, Sven and Ernst, Ernst and Esser, Frank and Büchel, Florin},
  date = {2017-08-03},
  journaltitle = {Information, Communication \& Society},
  volume = {20},
  number = {8},
  pages = {1109--1126},
  publisher = {Routledge},
  issn = {1369-118X},
  doi = {10.1080/1369118X.2016.1207697}
}

@article{engesserPopulistOnlineCommunication2017,
  title = {Populist Online Communication: Introduction to the Special Issue},
  shorttitle = {Populist Online Communication},
  author = {Engesser, Sven and Fawzi, Nayla and Larsson, Anders Olof},
  date = {2017-09-02},
  journaltitle = {Information, Communication \& Society},
  volume = {20},
  number = {9},
  pages = {1279--1292},
  publisher = {Routledge},
  issn = {1369-118X},
  doi = {10.1080/1369118X.2017.1328525}
}

@article{enliPersonalizedCampaignsPartyCentred2013,
  title = {Personalized {{Campaigns}} in {{Party-Centred Politics}}: {{Twitter}} and {{Facebook}} as Arenas for Political Communication},
  shorttitle = {Personalized {{Campaigns}} in {{Party-Centred Politics}}},
  author = {Enli, Gunn Sara and family=Skogerbø, given=Eli, prefix=and, useprefix=true},
  date = {2013-06-01},
  journaltitle = {Information, Communication \& Society},
  volume = {16},
  number = {5},
  pages = {757--774},
  publisher = {Routledge},
  issn = {1369-118X},
  doi = {10.1080/1369118X.2013.782330}
}

@incollection{esserComparingNewsNational2012,
  title = {Comparing {{News}} on {{National Elections}}},
  booktitle = {The {{Handbook}} of {{Comparative Communication Research}}},
  author = {Esser, Frank and Strömbäck, Jesper},
  date = {2012},
  publisher = {Routledge},
  isbn = {978-0-203-14910-2},
  pagetotal = {19}
}

@book{esserComparingPoliticalCommunication2004,
  title = {Comparing {{Political Communication}}: {{Theories}}, {{Cases}}, and {{Challenges}}},
  shorttitle = {Comparing {{Political Communication}}},
  editor = {Esser, Frank and Pfetsch, Barbara},
  date = {2004},
  series = {Communication, {{Society}} and {{Politics}}},
  publisher = {Cambridge University Press},
  location = {Cambridge},
  doi = {10.1017/CBO9780511606991},
  isbn = {978-0-521-82831-4}
}

@article{fridkinVariabilityCitizensReactions2011,
  title = {Variability in {{Citizens}}’ {{Reactions}} to {{Different Types}} of {{Negative Campaigns}}},
  author = {Fridkin, Kim L. and Kenney, Patrick},
  date = {2011},
  journaltitle = {American Journal of Political Science},
  volume = {55},
  number = {2},
  pages = {307--325},
  issn = {1540-5907},
  doi = {10.1111/j.1540-5907.2010.00494.x},
  langid = {english}
}

@book{geerDefenseNegativityAttack2006,
  title = {In {{Defense}} of {{Negativity}}: {{Attack Ads}} in {{Presidential Campaigns}}},
  author = {Geer, John G.},
  date = {2006-04-01},
  publisher = {University of Chicago Press},
  doi = {10.7208/chicago/9780226285009.001.0001},
  isbn = {978-0-226-28498-9}
}

@article{gilardiChatGPTOutperformsCrowd2023,
  title = {{{ChatGPT}} Outperforms Crowd Workers for Text-Annotation Tasks},
  author = {Gilardi, Fabrizio and Alizadeh, Meysam and Kubli, Maël},
  date = {2023-07-25},
  journaltitle = {Proceedings of the National Academy of Sciences},
  volume = {120},
  number = {30},
  pages = {e2305016120},
  publisher = {Proceedings of the National Academy of Sciences},
  doi = {10.1073/pnas.2305016120}
}

@article{grimmerTextDataPromise2013,
  title = {Text as {{Data}}: {{The Promise}} and {{Pitfalls}} of {{Automatic Content Analysis Methods}} for {{Political Texts}}},
  shorttitle = {Text as {{Data}}},
  author = {Grimmer, Justin and Stewart, Brandon M.},
  date = {2013-07},
  journaltitle = {Political Analysis},
  volume = {21},
  number = {3},
  pages = {267--297},
  issn = {1047-1987, 1476-4989},
  doi = {10.1093/pan/mps028},
  langid = {english}
}

@article{hansenNegativeCampaigningMultiparty2008,
  title = {Negative {{Campaigning}} in a {{Multiparty System}}},
  author = {Hansen, Kasper M. and Pedersen, Rasmus Tue},
  date = {2008},
  journaltitle = {Scandinavian Political Studies},
  volume = {31},
  number = {4},
  pages = {408--427},
  issn = {1467-9477},
  doi = {10.1111/j.1467-9477.2008.00213.x},
  langid = {english}
}

@article{haselmayerNegativeCampaigningIts2019,
  title = {Negative Campaigning and Its Consequences: A Review and a Look Ahead},
  shorttitle = {Negative Campaigning and Its Consequences},
  author = {Haselmayer, Martin},
  date = {2019-09-01},
  journaltitle = {French Politics},
  shortjournal = {Fr Polit},
  volume = {17},
  number = {3},
  pages = {355--372},
  issn = {1476-3427},
  doi = {10.1057/s41253-019-00084-8},
  langid = {english}
}

@article{haselmayerSentimentAnalysisPolitical2017,
  title = {Sentiment Analysis of Political Communication: Combining a Dictionary Approach with Crowdcoding},
  shorttitle = {Sentiment Analysis of Political Communication},
  author = {Haselmayer, Martin and Jenny, Marcelo},
  date = {2017-11-01},
  journaltitle = {Quality \& Quantity},
  shortjournal = {Qual Quant},
  volume = {51},
  number = {6},
  pages = {2623--2646},
  issn = {1573-7845},
  doi = {10.1007/s11135-016-0412-4},
  langid = {english}
}

@article{haynesAttackPoliticsPresidential1998,
  title = {Attack {{Politics}} in {{Presidential Nomination Campaigns}}: {{An Examination}} of the {{Frequency}} and {{Determinants}} of {{Intermediated Negative Messages Against Opponents}}},
  shorttitle = {Attack {{Politics}} in {{Presidential Nomination Campaigns}}},
  author = {Haynes, Audrey A. and Rhine, Staci L.},
  date = {1998-09-01},
  journaltitle = {Political Research Quarterly},
  volume = {51},
  number = {3},
  pages = {691--721},
  publisher = {SAGE Publications Inc},
  issn = {1065-9129},
  doi = {10.1177/106591299805100307},
  langid = {english}
}

@book{hixPoliticalPartiesEuropean1997,
  title = {Political {{Parties}} in the {{European Union}}},
  author = {Hix, Simon and Lord, Christopher},
  date = {1997},
  publisher = {Macmillan Education UK},
  location = {London},
  doi = {10.1007/978-1-349-25560-3},
  langid = {english}
}

@article{jollyChapelHillExpert2022a,
  title = {Chapel {{Hill Expert Survey}} Trend File, 1999–2019},
  author = {Jolly, Seth and Bakker, Ryan and Hooghe, Liesbet and Marks, Gary and Polk, Jonathan and Rovny, Jan and Steenbergen, Marco and Vachudova, Milada Anna},
  date = {2022-02-01},
  journaltitle = {Electoral Studies},
  shortjournal = {Electoral Studies},
  volume = {75},
  pages = {102420},
  issn = {0261-3794},
  doi = {10.1016/j.electstud.2021.102420}
}

@article{jungherrTwitterUseElection2016,
  title = {Twitter Use in Election Campaigns: {{A}} Systematic Literature Review},
  shorttitle = {Twitter Use in Election Campaigns},
  author = {Jungherr, Andreas},
  date = {2016-01-02},
  journaltitle = {Journal of Information Technology \& Politics},
  volume = {13},
  number = {1},
  pages = {72--91},
  publisher = {Routledge},
  issn = {1933-1681},
  doi = {10.1080/19331681.2015.1132401}
}

@article{keucheniusWhyItImportant2021,
  title = {Why It Is Important to Consider Negative Ties When Studying Polarized Debates: {{A}} Signed Network Analysis of a {{Dutch}} Cultural Controversy on {{Twitter}}},
  shorttitle = {Why It Is Important to Consider Negative Ties When Studying Polarized Debates},
  author = {Keuchenius, Anna and Törnberg, Petter and Uitermark, Justus},
  date = {2021-08-31},
  journaltitle = {PLOS ONE},
  shortjournal = {PLOS ONE},
  volume = {16},
  number = {8},
  pages = {e0256696},
  publisher = {Public Library of Science},
  issn = {1932-6203},
  doi = {10.1371/journal.pone.0256696},
  langid = {english}
}

@article{klingerAreCampaignsGetting2023,
  title = {Are {{Campaigns Getting Uglier}}, and {{Who Is}} to {{Blame}}? {{Negativity}}, {{Dramatization}} and {{Populism}} on {{Facebook}} in the 2014 and 2019 {{EP Election Campaigns}}},
  shorttitle = {Are {{Campaigns Getting Uglier}}, and {{Who Is}} to {{Blame}}?},
  author = {Klinger, Ulrike and Koc-Michalska, Karolina and Russmann, Uta},
  date = {2023-05-04},
  journaltitle = {Political Communication},
  volume = {40},
  number = {3},
  pages = {263--282},
  publisher = {Routledge},
  issn = {1058-4609},
  doi = {10.1080/10584609.2022.2133198}
}

@online{kojimaLargeLanguageModels2023,
  title = {Large {{Language Models}} Are {{Zero-Shot Reasoners}}},
  author = {Kojima, Takeshi and Gu, Shixiang Shane and Reid, Machel and Matsuo, Yutaka and Iwasawa, Yusuke},
  date = {2023-01-29},
  eprint = {2205.11916},
  eprinttype = {arXiv},
  eprintclass = {cs},
  doi = {10.48550/arXiv.2205.11916},
  pubstate = {prepublished}
}

@book{krippendorffContentAnalysisIntroduction2019,
  title = {Content {{Analysis}}: {{An Introduction}} to {{Its Methodology}}},
  shorttitle = {Content {{Analysis}}},
  author = {Krippendorff, Klaus},
  date = {2019},
  publisher = {SAGE Publications, Inc.},
  doi = {10.4135/9781071878781},
  isbn = {978-1-0718-7878-1},
  langid = {english}
}

@online{kristensen-mclachlanAreChatbotsReliable2025,
  title = {Are {{Chatbots Reliable Text Annotators}}? {{Sometimes}}},
  shorttitle = {Are {{Chatbots Reliable Text Annotators}}?},
  author = {Kristensen-McLachlan, Ross Deans and Canavan, Miceal and Kardos, Márton and Jacobsen, Mia and Aarøe, Lene},
  date = {2025-02-25},
  eprint = {2311.05769},
  eprinttype = {arXiv},
  eprintclass = {cs},
  doi = {10.48550/arXiv.2311.05769},
  pubstate = {prepublished}
}

@article{lauEffectMediaEnvironment2017,
  title = {Effect of {{Media Environment Diversity}} and {{Advertising Tone}} on {{Information Search}}, {{Selective Exposure}}, and {{Affective Polarization}}},
  author = {Lau, Richard R. and Andersen, David J. and Ditonto, Tessa M. and Kleinberg, Mona S. and Redlawsk, David P.},
  date = {2017-03-01},
  journaltitle = {Political Behavior},
  shortjournal = {Polit Behav},
  volume = {39},
  number = {1},
  pages = {231--255},
  issn = {1573-6687},
  doi = {10.1007/s11109-016-9354-8},
  langid = {english}
}

@article{lauEffectsNegativePolitical2007,
  title = {The {{Effects}} of {{Negative Political Campaigns}}: {{A Meta-Analytic Reassessment}}},
  shorttitle = {The {{Effects}} of {{Negative Political Campaigns}}},
  author = {Lau, Richard R. and Sigelman, Lee and Rovner, Ivy Brown},
  date = {2007-11},
  journaltitle = {The Journal of Politics},
  volume = {69},
  number = {4},
  pages = {1176--1209},
  publisher = {The University of Chicago Press},
  issn = {0022-3816},
  doi = {10.1111/j.1468-2508.2007.00618.x}
}

@book{lauNegativeCampaigningAnalysis2004a,
  title = {Negative Campaigning: An Analysis of {{U}}.{{S}}. {{Senate}} Elections},
  shorttitle = {Negative Campaigning},
  author = {Lau, Richard R. and Pomper, Gerald M.},
  date = {2004},
  series = {Campaigning {{American}} Style},
  publisher = {Rowman \& Littlefield},
  location = {Lanham, Md.},
  langid = {english},
  pagetotal = {177}
}

@article{lauNegativeCampaigningUS2001,
  title = {Negative {{Campaigning}} by {{US Senate Candidates}}},
  author = {Lau, Richard R. and Pomper, Gerald M.},
  date = {2001-01-01},
  journaltitle = {Party Politics},
  volume = {7},
  number = {1},
  pages = {69--87},
  publisher = {SAGE Publications Ltd},
  issn = {1354-0688},
  doi = {10.1177/1354068801007001004},
  langid = {english}
}

@article{lichtMeasuringUnderstandingParties2024,
  title = {Measuring and {{Understanding Parties}}’ {{Anti-elite Strategies}}},
  author = {Licht, Hauke and Abou-Chadi, Tarik and Barberá, Pablo and Hua, Whitney},
  date = {2024-04-03},
  journaltitle = {The Journal of Politics},
  pages = {000--000},
  publisher = {The University of Chicago Press},
  issn = {0022-3816},
  doi = {10.1086/730711}
}

@article{lindstadtWhenExpertsDisagree2020,
  title = {When {{Experts Disagree}}: {{Response Aggregation}} and Its {{Consequences}} in {{Expert Surveys}}},
  shorttitle = {When {{Experts Disagree}}},
  author = {Lindstädt, René and Proksch, Sven-Oliver and Slapin, Jonathan B.},
  date = {2020-07},
  journaltitle = {Political Science Research and Methods},
  volume = {8},
  number = {3},
  pages = {580--588},
  issn = {2049-8470, 2049-8489},
  doi = {10.1017/psrm.2018.52},
  langid = {english}
}

@inproceedings{lucyGenderRepresentationBias2021,
  title = {Gender and {{Representation Bias}} in {{GPT-3 Generated Stories}}},
  booktitle = {Proceedings of the {{Third Workshop}} on {{Narrative Understanding}}},
  author = {Lucy, Li and Bamman, David},
  editor = {Akoury, Nader and Brahman, Faeze and Chaturvedi, Snigdha and Clark, Elizabeth and Iyyer, Mohit and Martin, Lara J.},
  date = {2021-06},
  pages = {48--55},
  publisher = {Association for Computational Linguistics},
  location = {Virtual},
  doi = {10.18653/v1/2021.nuse-1.5},
  eventtitle = {{{NUSE-WNU}} 2021}
}

@article{maierMappingDriversNegative2023,
  title = {Mapping the Drivers of Negative Campaigning: {{Insights}} from a Candidate Survey},
  shorttitle = {Mapping the Drivers of Negative Campaigning},
  author = {Maier, Jürgen and Nai, Alessandro},
  date = {2023-03-01},
  journaltitle = {International Political Science Review},
  volume = {44},
  number = {2},
  pages = {195--211},
  publisher = {SAGE Publications Ltd},
  issn = {0192-5121},
  doi = {10.1177/0192512121994512},
  langid = {english}
}

@article{maierWhenCandidatesAttack2017,
  title = {When Do Candidates Attack in Election Campaigns? {{Exploring}} the Determinants of Negative Candidate Messages in {{German}} Televised Debates},
  shorttitle = {When Do Candidates Attack in Election Campaigns?},
  author = {Maier, Jürgen and Jansen, Carolin},
  date = {2017-09-01},
  journaltitle = {Party Politics},
  volume = {23},
  number = {5},
  pages = {549--559},
  publisher = {SAGE Publications Ltd},
  issn = {1354-0688},
  doi = {10.1177/1354068815610966},
  langid = {english}
}

@article{maierWhenConflictFuels2022,
  title = {When Conflict Fuels Negativity. {{A}} Large-Scale Comparative Investigation of the Contextual Drivers of Negative Campaigning in Elections Worldwide},
  author = {Maier, Jürgen and Nai, Alessandro},
  date = {2022-04-01},
  journaltitle = {The Leadership Quarterly},
  shortjournal = {The Leadership Quarterly},
  volume = {33},
  number = {2},
  pages = {101564},
  issn = {1048-9843},
  doi = {10.1016/j.leaqua.2021.101564}
}

@article{martinDeepeningRiftNegative2024,
  title = {Deepening the Rift: {{Negative}} Campaigning Fosters Affective Polarization in Multiparty Elections},
  shorttitle = {Deepening the Rift},
  author = {Martin, Danielle and Nai, Alessandro},
  date = {2024-02-01},
  journaltitle = {Electoral Studies},
  shortjournal = {Electoral Studies},
  volume = {87},
  pages = {102745},
  issn = {0261-3794},
  doi = {10.1016/j.electstud.2024.102745}
}

@article{mendozaFleetingAllureDark2024,
  title = {The {{Fleeting Allure}} of {{Dark Campaigns}}: {{Backlash}} from {{Negative}} and {{Uncivil Campaigning}} in the {{Presence}} of ({{Better}}) {{Alternatives}}},
  shorttitle = {The {{Fleeting Allure}} of {{Dark Campaigns}}},
  author = {Mendoza, Philipp and Nai, Alessandro and Bos, Linda},
  date = {2024-09-02},
  journaltitle = {Political Communication},
  volume = {41},
  number = {5},
  pages = {693--718},
  publisher = {Routledge},
  issn = {1058-4609},
  doi = {10.1080/10584609.2024.2314604}
}

@article{mikhaylovCoderReliabilityMisclassification2012,
  title = {Coder {{Reliability}} and {{Misclassification}} in the {{Human Coding}} of {{Party Manifestos}}},
  author = {Mikhaylov, Slava and Laver, Michael and Benoit, Kenneth R.},
  date = {2012-01},
  journaltitle = {Political Analysis},
  volume = {20},
  number = {1},
  pages = {78--91},
  issn = {1047-1987, 1476-4989},
  doi = {10.1093/pan/mpr047},
  langid = {english}
}

@book{moffittGlobalRisePopulism2016,
  title = {The Global Rise of Populism: Performance, Political Style, and Representation},
  shorttitle = {The Global Rise of Populism},
  author = {Moffitt, Benjamin},
  date = {2016},
  publisher = {Stanford University Press},
  location = {Stanford, California},
  isbn = {978-0-8047-9613-2},
  pagetotal = {224}
}

@book{muddePopulistRadicalRight2007,
  title = {Populist {{Radical Right Parties}} in {{Europe}}},
  author = {Mudde, Cas},
  date = {2007},
  publisher = {Cambridge University Press},
  location = {Cambridge},
  doi = {10.1017/CBO9780511492037},
  isbn = {978-0-521-85081-0}
}

@article{muddePopulistZeitgeist2004,
  title = {The {{Populist Zeitgeist}}},
  author = {Mudde, Cas},
  date = {2004},
  journaltitle = {Government and Opposition},
  volume = {39},
  number = {4},
  pages = {541--563},
  issn = {1477-7053},
  doi = {10.1111/j.1477-7053.2004.00135.x},
  langid = {english}
}

@article{naiGoingNegativeWorldwide2020,
  title = {Going {{Negative}}, {{Worldwide}}: {{Towards}} a {{General Understanding}} of {{Determinants}} and {{Targets}} of {{Negative Campaigning}}},
  shorttitle = {Going {{Negative}}, {{Worldwide}}},
  author = {Nai, Alessandro},
  date = {2020-07},
  journaltitle = {Government and Opposition},
  volume = {55},
  number = {3},
  pages = {430--455},
  issn = {0017-257X, 1477-7053},
  doi = {10.1017/gov.2018.32},
  langid = {english}
}

@article{nosAnderePartijenNiet2024,
  entrysubtype = {newspaper},
  title = {Andere partijen niet blij met tweet Wilders, maar het zal hem 'worst zijn'},
  author = {{NOS}},
  date = {2024-04-18},
  langid = {dutch}
}

@online{ollionChatGPTTextAnnotation2023,
  title = {{{ChatGPT}} for {{Text Annotation}}? {{Mind}} the {{Hype}}!},
  shorttitle = {{{ChatGPT}} for {{Text Annotation}}?},
  author = {Ollion, Étienne and Shen, Rubing and Macanovic, Ana and Chatelain, Arnault},
  date = {2023-10-04},
  eprinttype = {OSF},
  doi = {10.31235/osf.io/x58kn},
  langid = {american},
  pubstate = {prepublished}
}

@article{oschatzThatsNotAppropriate2024,
  title = {"{{That}}'s {{Not Appropriate}}!" {{Examining Social Norms}} as {{Predictors}} of {{Negative Campaigning}}},
  author = {Oschatz, Corinna and Maier, Jürgen and Dian, Mona and Geber, Sarah},
  date = {2024-07-18},
  journaltitle = {Political Behavior},
  shortjournal = {Polit Behav},
  issn = {1573-6687},
  doi = {10.1007/s11109-024-09958-2},
  langid = {english}
}

@article{oschatzTwitterNewsAnalysis2022,
  title = {Twitter in the {{News}}: {{An Analysis}} of {{Embedded Tweets}} in {{Political News Coverage}}},
  shorttitle = {Twitter in the {{News}}},
  author = {Oschatz, Corinna and Stier, Sebastian and Maier, Jürgen},
  date = {2022-10-21},
  journaltitle = {Digital Journalism},
  volume = {10},
  number = {9},
  pages = {1526--1545},
  publisher = {Routledge},
  issn = {2167-0811},
  doi = {10.1080/21670811.2021.1912624}
}

@article{pappMacroLevelDrivingFactors2019,
  title = {The {{Macro-Level Driving Factors}} of {{Negative Campaigning}} in {{Europe}}},
  author = {Papp, Zsófia and Patkós, Veronika},
  date = {2019-01-01},
  journaltitle = {The International Journal of Press/Politics},
  volume = {24},
  number = {1},
  pages = {27--48},
  publisher = {SAGE Publications Inc},
  issn = {1940-1612},
  doi = {10.1177/1940161218803426},
  langid = {english}
}

@article{petkevicPoliticalAttacks2802022,
  title = {Political {{Attacks}} in 280 {{Characters}} or {{Less}}: {{A New Tool}} for the {{Automated Classification}} of {{Campaign Negativity}} on {{Social Media}}},
  shorttitle = {Political {{Attacks}} in 280 {{Characters}} or {{Less}}},
  author = {Petkevic, Vladislav and Nai, Alessandro},
  date = {2022-05-01},
  journaltitle = {American Politics Research},
  volume = {50},
  number = {3},
  pages = {279--302},
  publisher = {SAGE Publications Inc},
  issn = {1532-673X},
  doi = {10.1177/1532673X211055676},
  langid = {english}
}

@article{pinkletonEffectsNegativeComparative1997,
  title = {The {{Effects}} of {{Negative Comparative Political Advertising}} on {{Candidate Evaluations}} and {{Advertising Evaluations}}: {{An Exploration}}},
  shorttitle = {The {{Effects}} of {{Negative Comparative Political Advertising}} on {{Candidate Evaluations}} and {{Advertising Evaluations}}},
  author = {Pinkleton, Bruce},
  date = {1997-03-01},
  journaltitle = {Journal of Advertising},
  volume = {26},
  number = {1},
  pages = {19--29},
  publisher = {Routledge},
  issn = {0091-3367},
  doi = {10.1080/00913367.1997.10673515}
}

@online{raffelExploringLimitsTransfer2023,
  title = {Exploring the {{Limits}} of {{Transfer Learning}} with a {{Unified Text-to-Text Transformer}}},
  author = {Raffel, Colin and Shazeer, Noam and Roberts, Adam and Lee, Katherine and Narang, Sharan and Matena, Michael and Zhou, Yanqi and Li, Wei and Liu, Peter J.},
  date = {2023-09-19},
  eprint = {1910.10683},
  eprinttype = {arXiv},
  eprintclass = {cs},
  doi = {10.48550/arXiv.1910.10683},
  pubstate = {prepublished}
}

@article{rathjeGPTEffectiveTool2024,
  title = {{{GPT}} Is an Effective Tool for Multilingual Psychological Text Analysis},
  author = {Rathje, Steve and Mirea, Dan-Mircea and Sucholutsky, Ilia and Marjieh, Raja and Robertson, Claire E. and Van Bavel, Jay J.},
  date = {2024-08-20},
  journaltitle = {Proceedings of the National Academy of Sciences},
  volume = {121},
  number = {34},
  pages = {e2308950121},
  publisher = {Proceedings of the National Academy of Sciences},
  doi = {10.1073/pnas.2308950121}
}

@article{roeseBacklashEffectsAttack1993,
  title = {Backlash {{Effects}} in {{Attack Politics}}},
  author = {Roese, Neal J. and Sande, Gerald N.},
  date = {1993},
  journaltitle = {Journal of Applied Social Psychology},
  volume = {23},
  number = {8},
  pages = {632--653},
  issn = {1559-1816},
  doi = {10.1111/j.1559-1816.1993.tb01106.x},
  langid = {english}
}

@online{tornbergBestPracticesText2024,
  title = {Best {{Practices}} for {{Text Annotation}} with {{Large Language Models}}},
  author = {Törnberg, Petter},
  date = {2024-02-05},
  eprint = {2402.05129},
  eprinttype = {arXiv},
  eprintclass = {cs},
  doi = {10.48550/arXiv.2402.05129},
  pubstate = {prepublished}
}

@article{tornbergLargeLanguageModels2024,
  title = {Large {{Language Models Outperform Expert Coders}} and {{Supervised Classifiers}} at {{Annotating Political Social Media Messages}}},
  author = {Törnberg, Petter},
  date = {2024-09-22},
  journaltitle = {Social Science Computer Review},
  pages = {08944393241286471},
  publisher = {SAGE Publications Inc},
  issn = {0894-4393},
  doi = {10.1177/08944393241286471},
  langid = {english}
}

@article{tornbergWhenPartiesLie2025,
  title = {When {{Do Parties Lie}}? {{Misinformation}} and {{Radical-Right Populism Across}} 26 {{Countries}}},
  shorttitle = {When {{Do Parties Lie}}?},
  author = {Törnberg, Petter and Chueri, Juliana},
  date = {2025-01-13},
  journaltitle = {The International Journal of Press/Politics},
  pages = {19401612241311886},
  publisher = {SAGE Publications Inc},
  issn = {1940-1612},
  doi = {10.1177/19401612241311886},
  langid = {english}
}

@article{vanatteveldtValiditySentimentAnalysis2021,
  title = {The {{Validity}} of {{Sentiment Analysis}}: {{Comparing Manual Annotation}}, {{Crowd-Coding}}, {{Dictionary Approaches}}, and {{Machine Learning Algorithms}}},
  shorttitle = {The {{Validity}} of {{Sentiment Analysis}}},
  author = {family=Atteveldt, given=Wouter, prefix=van, useprefix=true and family=Velden, given=Mariken A. C. G., prefix=van der, useprefix=true and Boukes, Mark},
  date = {2021-04-03},
  journaltitle = {Communication Methods and Measures},
  volume = {15},
  number = {2},
  pages = {121--140},
  publisher = {Routledge},
  issn = {1931-2458},
  doi = {10.1080/19312458.2020.1869198}
}

@article{vanvlietTwitterParliamentarianDatabase2020,
  title = {The {{Twitter}} Parliamentarian Database: {{Analyzing Twitter}} Politics across 26 Countries},
  shorttitle = {The {{Twitter}} Parliamentarian Database},
  author = {family=Vliet, given=Livia, prefix=van, useprefix=true and Törnberg, Petter and Uitermark, Justus},
  date = {2020-09-16},
  journaltitle = {PLOS ONE},
  shortjournal = {PLOS ONE},
  volume = {15},
  number = {9},
  pages = {e0237073},
  publisher = {Public Library of Science},
  issn = {1932-6203},
  doi = {10.1371/journal.pone.0237073},
  langid = {english}
}

@article{volkensStrengthsWeaknessesApproaches2007,
  title = {Strengths and Weaknesses of Approaches to Measuring Policy Positions of Parties},
  author = {Volkens, Andrea},
  date = {2007-03-01},
  journaltitle = {Electoral Studies},
  shortjournal = {Electoral Studies},
  volume = {26},
  number = {1},
  pages = {108--120},
  issn = {0261-3794},
  doi = {10.1016/j.electstud.2006.04.003}
}

@incollection{walterExplainingUseAttack2015,
  title = {Explaining the {{Use}} of {{Attack Behaviour}} in the {{Electoral Battlefield}}: {{A Literature Overview}}},
  booktitle = {New Perspectives on Negative Campaigning: Why Attack Politics Matters},
  author = {Walter, Annemarie S. and Nai, Alessandro},
  editor = {Nai, Alessandro and Walter, Annemarie S.},
  date = {2015},
  series = {{{ECPR}} - {{Studies}} in {{European}} Political Science},
  publisher = {ECPR Press},
  location = {Colchester, UK},
  isbn = {978-1-78552-128-7}
}

@article{walterUnintendedConsequencesNegative2019,
  title = {Unintended Consequences of Negative Campaigning: {{Backlash}} and Second-Preference Boost Effects in a Multi-Party Context},
  shorttitle = {Unintended Consequences of Negative Campaigning},
  author = {Walter, Annemarie S. and family=Eijk, given=Cees, prefix=van der, useprefix=true},
  date = {2019-08-01},
  journaltitle = {The British Journal of Politics and International Relations},
  volume = {21},
  number = {3},
  pages = {612--629},
  publisher = {SAGE Publications},
  issn = {1369-1481},
  doi = {10.1177/1369148119842038},
  langid = {english}
}

@article{walterWhenGlovesCome2013,
  title = {When the Gloves Come off: {{Inter-party}} Variation in Negative Campaigning in {{Dutch}} Elections, 1981–2010},
  shorttitle = {When the Gloves Come Off},
  author = {Walter, Annemarie S. and family=Brug, given=Wouter, prefix=van der, useprefix=true},
  date = {2013-10-01},
  journaltitle = {Acta Politica},
  shortjournal = {Acta Polit},
  volume = {48},
  number = {4},
  pages = {367--388},
  issn = {1741-1416},
  doi = {10.1057/ap.2013.5},
  langid = {english}
}

@article{walterWhenStakesAre2014,
  title = {When the {{Stakes Are High}}: {{Party Competition}} and {{Negative Campaigning}}},
  shorttitle = {When the {{Stakes Are High}}},
  author = {Walter, Annemarie S. and family=Brug, given=Wouter, prefix=van der, useprefix=true and family=Praag, given=Philip, prefix=van, useprefix=true},
  date = {2014-03-01},
  journaltitle = {Comparative Political Studies},
  volume = {47},
  number = {4},
  pages = {550--573},
  publisher = {SAGE Publications Inc},
  issn = {0010-4140},
  doi = {10.1177/0010414013488543},
  langid = {english}
}

@article{widmannCreatingComparingDictionary2023,
  title = {Creating and {{Comparing Dictionary}}, {{Word Embedding}}, and {{Transformer-Based Models}} to {{Measure Discrete Emotions}} in {{German Political Text}}},
  author = {Widmann, Tobias and Wich, Maximilian},
  date = {2023-10},
  journaltitle = {Political Analysis},
  volume = {31},
  number = {4},
  pages = {626--641},
  issn = {1047-1987, 1476-4989},
  doi = {10.1017/pan.2022.15},
  langid = {english}
}

@online{yuOpenClosedSmall2023,
  title = {Open, {{Closed}}, or {{Small Language Models}} for {{Text Classification}}?},
  author = {Yu, Hao and Yang, Zachary and Pelrine, Kellin and Godbout, Jean Francois and Rabbany, Reihaneh},
  date = {2023-08-19},
  eprint = {2308.10092},
  eprinttype = {arXiv},
  eprintclass = {cs},
  doi = {10.48550/arXiv.2308.10092},
  pubstate = {prepublished}
}

@article{meijersMeasuringPopulismPolitical2021,
  title = {Measuring {{Populism}} in {{Political Parties}}: {{Appraisal}} of a {{New Approach}}},
  shorttitle = {Measuring {{Populism}} in {{Political Parties}}},
  author = {Meijers, Maurits J. and Zaslove, Andrej},
  date = {2021-02-01},
  journaltitle = {Comparative Political Studies},
  volume = {54},
  number = {2},
  pages = {372--407},
  publisher = {SAGE Publications Inc},
  issn = {0010-4140},
  doi = {10.1177/0010414020938081},
  langid = {english}
}

@article{zasloveStatePopulismIntroducing2025,
  title = {The State of Populism: {{Introducing}} the 2023 Wave of the {{Populism}} and Political Parties Expert Survey},
  shorttitle = {The State of Populism},
  author = {Zaslove, Andrej and Huber, Robert A and Meijers, Maurits J.},
  date = {2025-07-30},
  journaltitle = {Party Politics},
  pages = {13540688251361813},
  publisher = {SAGE Publications Ltd},
  issn = {1354-0688},
  doi = {10.1177/13540688251361813},
  langid = {english}
}

\end{document}